# Developing Approaches for Solving a Telecommunications Feature Subscription Problem

**David Lesaint**                                          DAVID.LESAINT@BT.COM
*Business Modelling and Operational Transformation,*
*BT Research,*
*UK.*

**Deepak Mehta**                                             D.MEHTA@4C.UCC.IE
**Barry O'Sullivan**                                     B.OSULLIVAN@4C.UCC.IE
**Luis Quesada**                                           L.QUESADA@4C.UCC.IE
**Nic Wilson**                                              N.WILSON@4C.UCC.IE
*Cork Constraint Computation Centre,*
*University College Cork,*
*Ireland.*

## Abstract

Call control features (e.g., call-divert, voice-mail) are primitive options to which users can subscribe off-line to personalise their service. The configuration of a feature subscription involves choosing and sequencing features from a catalogue and is subject to constraints that prevent undesirable feature interactions at run-time. When the subscription requested by a user is inconsistent, one problem is to find an optimal relaxation, which is a generalisation of the feedback vertex set problem on directed graphs, and thus it is an NP-hard task. We present several constraint programming formulations of the problem. We also present formulations using partial weighted maximum Boolean satisfiability and mixed integer linear programming. We study all these formulations by experimentally comparing them on a variety of randomly generated instances of the feature subscription problem.

## 1. Introduction

Information and communication services, from news feeds to internet telephony, are playing an increasing, and potentially disruptive, role in our daily lives. As a result, service providers seek to develop personalisation solutions allowing customers to control and enrich their service. In telephony, for instance, personalisation relies on the provisioning of call control features. A feature is an increment of functionality which, if activated, modifies the basic service behaviour in systematic or non-systematic ways, e.g., do-not-disturb, multi-media ring-back tones, call-divert-on-busy, credit-card-calling.

Modern service delivery platforms provide the ability to implement features as modular applications and compose them "on demand" when setting up live sessions, that is, consistently with the feature subscriptions preconfigured by participants. The architectural style commonly found in platforms that are based on the Session Initiation Protocol (Rosenberg, Schulzrinne, Camarillo, Johnston, Peterson, Sparks, Handley, & Schooler, 2002; Sparks, 2007) notably, the Internet Multimedia Subsystem (Poikselka, Mayer, Khartabil, & Niemi, 2006), consists of chaining applications between end-points. In this context, a personali-sation approach consists of exposing catalogues of call-control features to subscribers and letting them select and sequence the features of their choice.

Not all sequences of features are acceptable, however, due to the possible occurrence of feature interactions (Calder, Kolberg, Magill, & Reiff-Marganiec, 2003). A feature interaction is "some way in which a feature modifies or influences the behaviour of another feature in generating the system's overall behaviour" (Bond, Cheung, Purdy, Zave, & Ramming, 2004). For instance, a do-not-disturb feature will block any incoming call and cancel the effect of any subsequent feature subscribed by the callee. This is an undesirable interaction: as shown in Figure 1, the call originating from caller X will never reach call-logging feature of callee Y. However, if call-logging is placed before do-not-disturb then both features will play their role.

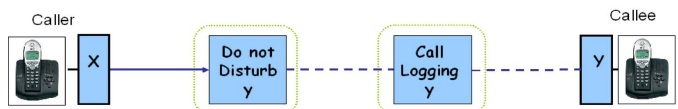

Figure 1: An example of an undesirable feature interaction.

Distributed Feature Composition (DFC) provides a method and a formal architecture to address feature interactions (Jackson & Zave, 1998, 2003; Bond et al., 2004). The method consists of constraining the selection and sequencing of features by prescribing constraints that prevent undesirable interactions. These feature interaction resolution constraints are represented in a catalogue as precedence or exclusion constraints. A precedence constraint, $i \prec j$, means that if the features $i$ and $j$ are part of the same sequence then $i$ must precede $j$. An exclusion constraint between $i$ and $j$ means that they cannot be together in any sequence. Note that an exclusion constraint between $i$ and $j$ can be expressed as a pair of two precedence constraints $i \prec j$ and $j \prec i$. Undesirable interactions are then avoided by rejecting any sequence that does not satisfy the catalogue precedence constraints.

Informally, a *feature subscription* is defined by a set of features, a set of precedence constraints specified by a user and a set of precedence constraints prescribed by the feature catalogue. The task is to find a sequence of the user-selected features subject to the catalogue precedence constraints and the user-specified precedence constraints. It may not always be possible to construct such a sequence, in which case the task is to find a relaxation of the feature subscription that is consistent and closest to the initial requirements of the user (Lesaint, Mehta, O'Sullivan, Quesada, & Wilson, 2008b). In this paper, we show that checking the consistency of a feature subscription is polynomial in time, but finding an optimal relaxation of a subscription, when inconsistent, is NP-hard.

We present several formulations of finding an optimal relaxation of a feature subscription using *constraint programming*. We present a simple constraint optimisation problem formulation of our problem and investigate the impact of maintaining three different levels of consistency on decision variables within depth-first branch and bound. The first one is arc consistency (Rossi, van Beek, & Walsh, 2006a), which is commonly used. The second is singleton arc consistency and the third is restricted singleton arc consistency (RSAC). We also present a formulation of our problem based on a soft global constraint, which we call SoftPrec (Lesaint, Mehta, O'Sullivan, Quesada, & Wilson, 2009). We further present a formulation based on the weighted constraint satisfaction problem framework (Rossi, van Beek, & Walsh, 2006b). We also consider *partial weighted maximum satisfiability* (Biere, Heule, van Maaren, & Walsh, 2009), and *mixed integer linear programming*. We present the formulations using these approaches and discuss their differences with respect to the constraint programming formulations.

Notice that finding an optimal relaxation of a feature subscription is a generalisation of the well-known *feedback vertex set problem* as well as the *feedback arc set problem* (Garey & Johnson, 1979). Given a directed graph $G = \langle V, E \rangle$ with set of vertices $V$ and set of edges $E$, the feedback vertex (arc) set problem is to find a smallest $V' \subseteq V$ ($E' \subseteq E$) whose deletion makes the graph acyclic. Although in this paper we focus only on a particular telecommunication problem, the techniques studied here are also applicable to other domains where the feedback vertex/arc set problem is encountered, e.g., circuit design, deadlock prevention, VLSI testing, stabilization of synchronous systems (Festa, Pardalos, & Resende, 1999, Section 5). There are also applications in chemistry when it comes to sorting a list of samples of complex mixtures according to their compositions in the presence of missing data, i.e., when not all components are measured in all samples (Fried, Hordijk, Prohaska, Stadler, & Stadler, 2004).

The remainder of this paper is organised as follows. Section 2 presents the necessary background required for this paper. We introduce the notion of feature subscription in Section 3. In Section 4 we reformulate the original problem in order to relate it more easily to well-known problems existing in the literature. In Section 5 we present an algorithm for dealing with symmetries introduced when the original subscription is reformulated. We introduce the notion of relaxation of an inconsistent subscription in Section 6 and prove that finding an optimal relaxation of an inconsistent subscription is NP-Hard. In Section 7 we model the problem of finding such an optimal relaxation as a constraint optimisation problem. In Section 8, we present two other constraint programming approaches based on the notions of global constraints and weighted constraint satisfaction problems. In Sections 9 and 10, the partial weighted maximum satisfiability and mixed integer linear programming formulations of the problem are described. The empirical evaluation of all these approaches is shown in Section 11. Finally our conclusions and future directions are presented in Section 12.

## 2. Background

In this section we present a set of concepts on binary relations and constraint programming that will be used in the next sections.

### 2.1 Binary Relations

A binary relation over a finite set $X$ is an association of elements of $X$ with elements of $X$. Let $R$ be a binary relation over a finite set $X$. A relation $R$ on a set $X$ is *irreflexive* if and only if there is no $x \in X$ such that $\langle x, x \rangle \in R$. A relation $R$ on a set $X$ is *transitive* if and only if for all $x$, $y$ and $z$ in $X$, $[\langle x, y \rangle \in R] \wedge [\langle y, z \rangle \in R] \Rightarrow [\langle x, z \rangle \in R]$. The *transitive closure* of a binary relation $R$ on a set $X$ is the smallest transitive relation on $X$ that contains $R$. We use the notation $R^*$ to denote the transitive closure of $R$. A relation $R$ on a set $X$ is *asymmetric* if and only if for all $x$, $y$ in $X$, $[\langle x, y \rangle \in R] \Rightarrow [\langle y, x \rangle \notin R]$. A relation $R$ on a set $X$ is total if and only if for any $x$, $y$ in $X$, either $\langle x, y \rangle \in R$ or $\langle y, x \rangle \in R$. A *strict partial order* is a binary relation that is irreflexive and transitive. A *strict total order* is a binary relation that is transitive, asymmetric and total. The *transpose* of a relation $R$, denoted $\widehat{R}$, is the set $\{\langle y, x \rangle | \langle x, y \rangle \in R\}$. The *restriction* of $R$ on the set $Y$, denoted $R{\downarrow}_Y$, is the set $\{\langle x, y \rangle \in R | \{x, y\} \subseteq Y\}$. Any binary relation $R$ on set $X$ can also be viewed as a directed graph where the nodes correspond to the elements in $X$ and ordered pairs in $R$ correspond to the edges of the graph.

## 2.2 Constraint Programming

Constraint Programming (CP) has been successfully used in many applications such as planning, scheduling, resource allocation, routing, and bio-informatics (Wallace, 1996). Problems are primarily stated as a Constraint Satisfaction Problems (CSPs), that is a finite set of variables with finite domains, together with a finite set of constraints. A solution of a CSP is an assignment of a value to each variable such that all constraints are satisfied simultaneously. The basic approach for solving a CSP instance is to use a backtracking search algorithm that interleaves two processes: *constraint propagation* and *labelling*. Constraint propagation helps in pruning values that cannot lead to a solution of the problem. Labelling involves assigning values to variables that may lead to a solution.

A binary constraint is said to be arc consistent if for every value in the domain of every variable, there exists a value in the domain of the other such that the pair of values satisfies the constraint between the variables. A non-binary constraint is generalised arc consistent if and only if for any value for a variable in its scope, there exists a value for every other variable in the scope such that the tuple satisfies the constraint (Rossi et al., 2006a). A CSP is said to be *Arc Consistent* (AC) if all its constraints are (generalised) arc consistent. A CSP is said to be *Singleton Arc Consistent* (SAC) if it has non-empty domains and for any assignment of a variable the resulting subproblem can be made AC (Bessiere, Stergiou, & Walsh, 2008). Mixed consistency means maintaining different levels of consistency on different variables of a problem. It has been shown that maintaining SAC on some variables and AC on the remaining variables of certain problems, such as job shop scheduling and radio link frequency assignment, can reduce the solution time (Lecoutre & Patrick, 2006).

Various generalisations of CSPs have been developed, where the objective is to find a solution that is optimal with respect to certain criteria such as costs, preferences or priorities. One of the most significant is the Constraint Optimisation Problem (COP). Here the goal is to find an optimal solution that either maximises or minimises an objective function depending upon the problem. The simplest COP formulation retains the CSP limitation of allowing only hard constraints but adds an objective function over the variables.

A depth-first branch and bound search algorithm is generally used to find a solution of a COP having an optimal value. In the case of maximisation, branch and bound search algorithm keeps the current optimal value of the solution while traversing the search tree. This value is a lower bound on the optimal value of the objective function. At each node of the search tree, the search algorithm computes an overestimation of the global value. This value is an upper bound on the best solution that extends the current partial solution. If the lower bound is greater than or equal to the upper bound, then a solution of a greater value than the current optimal value cannot be found below the current node, so the current branch is pruned and the algorithm backtracks.

## 3. Configuring Feature Subscriptions

In Distributed Feature Composition (DFC) each feature is implemented by one or more modules called Feature Box Types (FBT) and each FBT has many run-time instances called *feature boxes*. For simplicity, in this paper we assume that each feature is implemented by a single feature box and we associate features with feature boxes.

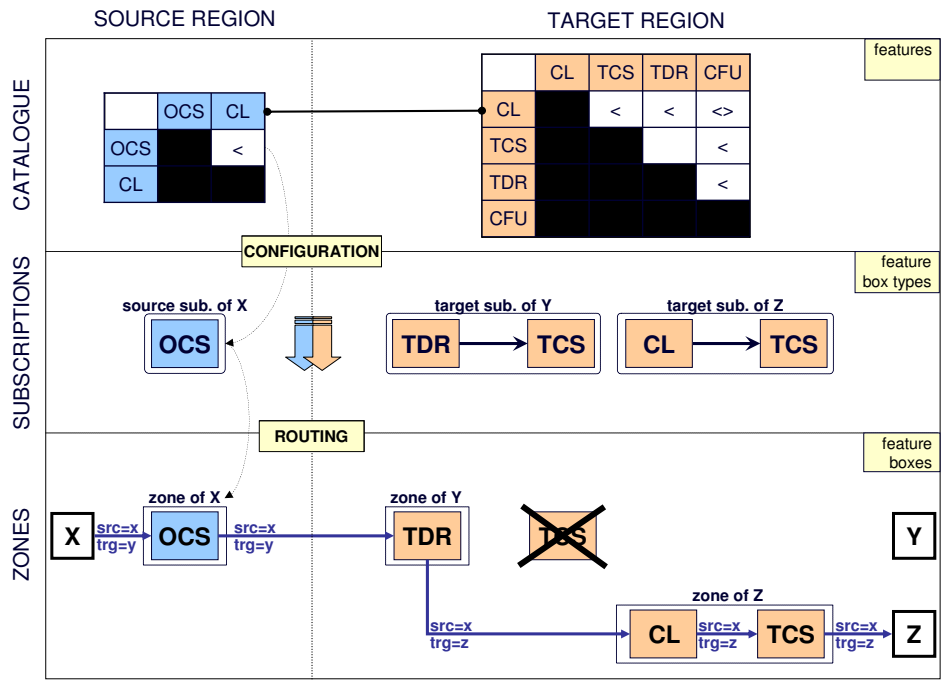

Figure 2: DFC: Catalogues, subscriptions and sessions.

DFC establishes a dialogue between endpoints by routing a set-up request encapsulating source and target addresses that are associated with source and target feature boxes respectively. Addresses may change along the way and DFC routers evolve the connection path accordingly. Starting from the feature box initiating the call, feature boxes are incorporated one after the other until a terminating box is reached. A router is used at each step to locate the next box and relay the set-up request. As shown in the third row of Figure 2, the routing method decomposes the connection path into a source and a target region and each region is further partitioned into *zones*. A source (target) zone is a sequence of feature boxes that execute for the same source (target) address. The first source zone is associated with the source address encapsulated in the initial set-up request, i.e, zone of $X$ in Figure 2. A change of source address in the source region, caused for instance by an identification feature, triggers the creation of a new source zone. If no such change occurs and the zone cannot be expanded further, routers switch to the target region. Likewise, a change of target address in the target region, as performed by Time-Dependent-Routing (TDR) in Figure 2, triggers the creation of a new target zone. If no such change occurs and the zone cannot be expanded further as for $Z$ in Figure 2, the request is sent to the final box identified by the encapsulated target address.

DFC routers are only concerned with locating feature boxes and assembling zones into regions. They do not make decisions as to the type and ordering of feature boxes appearing in a zone. They simply fetch this information from the pre-configured *feature subscription* that is associated with the address and region of the zone and use it to construct the zone. For instance, the zone of Z in Figure 2 results from the sequence of feature box types subscribed to by Z in the target region.

Subscriptions are pre-configured from the *feature catalogue* published by the service provider. The catalogue is a set of features. Features are classified as *source*, *target* or *reversible* (i.e., a subset of features that are both source and target) based on whether they can be subscribed to in the source region, the target region or both. For instance, the catalogue shown in the first row of Figure 2 includes Originating-Call-Screening (OCS) as a source feature, Terminating-Call-Screening (TCS), Time-Dependent-Routing (TDR), and Call-Forwarding-Unconditional (CFU) as target features, and Call-Logging (CL) as a reversible feature. A source feature is activated on behalf of a caller while a target feature is activated on behalf of a callee.

Constraints are formulated by designers on pairs of source features and pairs of target features to prevent undesirable feature interactions (Zave, 2003). A precedence constraint imposes a routing order between two features. The order is specified with respect to the direction of an outgoing call if the features are source (e.g., OCS must precede CL in Figure 2) and with respect to the direction of an incoming call if the features are target (e.g., CL must precede TCS). An exclusion constraint makes two features mutually exclusive, as for the case of CL and CFU in Figure 2. We encode an exclusion constraint between two features $f_i$ and $f_j$ as the pair of precedence constraints $f_i \prec f_j$ and $f_j \prec f_i$. For the sake of simplicity, we treat precedence constraints as ordered pairs, i.e., the precedence constraint $f_i \prec f_j$ is also viewed as $\langle f_i, f_j \rangle$.

**Definition 1** (Catalogue). *A catalogue is a tuple $\langle \mathcal{F}_s, \mathcal{H}_s, \mathcal{F}_t, \mathcal{H}_t \rangle$ where:*

- *$\mathcal{F}_s$ is the finite set of source features,*

- *$\mathcal{F}_t$ is the finite set of target features,*

- *$\mathcal{F}_s \cap \mathcal{F}_t$ is the finite set of reversible features,*

- *$\mathcal{H}_s$ is the set of source precedence constraints over $\mathcal{F}_s$, and*

- *$\mathcal{H}_t$ is the set of target precedence constraints over $\mathcal{F}_t$.*

The source (target) subscription associated with an address is a subset of source (target) catalogue features, a set of catalogue precedence constraints between source (target) features, and a set of user precedence constraints between source (target) features. For instance, the target subscription of Y shown in the second row of Figure 2 includes the target features TDR and TCS and the user precedence TDR $\prec$ TCS meaning that TDR should appear before TCS in the connection.

**Definition 2** (Feature Subscription). *Given a catalogue $\langle \mathcal{F}_s, \mathcal{H}_s, \mathcal{F}_t, \mathcal{H}_t \rangle$, a feature subscription is defined to be a pair of tuples $S_s = \langle F_s, H_s, P_s \rangle$ and $S_t = \langle F_t, H_t, P_t \rangle$ where:*

- *$F_s$ and $F_t$ are the user selected source and target features respectively such that $F_s \subseteq \mathcal{F}_s$, $F_t \subseteq \mathcal{F}_t$ and $F_s \cap F_t = F_t \cap \mathcal{F}_s$, i.e., any reversible feature in $F_s \cup F_t$ appears in both $F_s$ and $F_t$;*

- *$H_s$ is the set of source catalogue precedence constraints in $F_s$ given by $H_s = \mathcal{H}_s{\downarrow}_{F_s}$ $\cup \{(f \prec g) \in (F_s \cap F_t)^2 : g \prec f \in \mathcal{H}_t\}$;*

- *$H_t$ is the set of target catalogue precedence constraints in $F_t$ given by $H_t = \mathcal{H}_t{\downarrow}_{F_t}$ $\cup \{(f \prec g) \in (F_s \cap F_t)^2 : g \prec f \in \mathcal{H}_s\}$;*

- *$P_s$ is the set of source user precedence constraints over $F_s$, which satisfies $P_s \supseteq \{(f \prec g) \in (F_s \cap F_t)^2 : g \prec f \in P_t\}$;*

- *$P_t$ is the set of target user precedence constraints over $F_t$, which satisfies $P_t \supseteq \{(f \prec g) \in (F_s \cap F_t)^2 : g \prec f \in P_s\}$.*

Configuring a feature subscription involves selecting, parameterising and sequencing features in each region consistently with the catalogue constraints and other integrity rules (Jackson & Zave, 2003). In particular, the source and target regions of a subscription must include the same reversible features in inverse order, i.e. source and target regions are not configured independently.

**Definition 3** (Consistency of Feature Subscriptions). *We say that a feature subscription $S = \langle \langle F_s, H_s, P_s \rangle, \langle F_t, H_t, P_t \rangle \rangle$ is consistent if and only if there exists a strict total order $T_s$ on $F_s$ and a strict total order $T_t$ on $F_t$ such that*

1. *$T_s \supseteq H_s \cup P_s$*

2. *$T_t \supseteq H_t \cup P_t$*

3. *for all $f, g \in F_s \cap F_t, f \prec g \in T_s \Leftrightarrow g \prec f \in T_t$.*

The following configuration services may be provided to users submitting a feature subscription:

- (**verification**) Check the consistency of the subscription.

- (**filtering**) If the feature subscription is consistent, then compute its anti-subscription, i.e., the set of features and precedence constraints that would make it inconsistent if added.

- (**partial completion**) If the feature subscription is consistent, then compute the transitive closure of each region, i.e., $(H_s \cup P_s)^*$ and $(H_t \cup P_t)^*$.

- (**completion**) If the feature subscription is consistent, then compute a pair of strict total orders on source and target features such that points 1, 2 and 3 of Definition 3 are respected.

- (**relaxation**) If the feature subscription is inconsistent, then suggest consistent subscriptions obtained out of it by removing one more features or user precedences.

We formalise these tasks in the next section and describe their time complexities after reformulating the original definition of feature subscription.

## 4. Reformulating the Original Definition of Feature Subscription

By definition, a catalogue includes two sets of features and two sets of precedence constraints. In this section, we reformulate a catalogue by merging its source and target feature sets and by merging its source and target precedence sets. We transform feature subscriptions accordingly and show that the consistency of a subscription is equivalent to the acyclicity of its transformation. The new definitions are simpler and this reformulation allows us to establish relations with the other well-known problems existing in the literature.

The principle of the reformulation of a catalogue is to inverse and merge the target precedences with the source precedences. Specifically, a catalogue $\langle \mathcal{F}_s, \mathcal{H}_s, \mathcal{F}_t, \mathcal{H}_t \rangle$ is reformulated as $\langle \mathcal{F}_c, \mathcal{H}_c \rangle \equiv \langle \mathcal{F}_s \cup \mathcal{F}_t, \mathcal{H}_s \cup \widehat{\mathcal{H}_t} \rangle$, where $\widehat{\mathcal{H}_t}$ is the *transpose* of $\mathcal{H}_t$ such that $\forall \langle i, j \rangle \in \mathcal{F}_t{}^2 : \langle i, j \rangle \in \mathcal{H}_t \Leftrightarrow \langle j, i \rangle \in \widehat{\mathcal{H}_t}$. The definitions of (consistent) feature subscription are adapted as follows.

**Definition 4** (Feature Subscription). *A feature subscription $S$ of catalogue $\langle \mathcal{F}_c, \mathcal{H}_c \rangle$ is a tuple $\langle F, H, P \rangle$, where $F \subseteq \mathcal{F}_c$, $H = \mathcal{H}_c{\downarrow}_F$, and $P$ is a set of (user defined) precedence constraints on $F$.*

**Definition 5** (Consistency of the Reformulated Feature Subscription). *A feature subscription $\langle F, H, P \rangle$ of a catalogue $\langle \mathcal{F}_c, \mathcal{H}_c \rangle$ is defined to be consistent if and only if there exists a total order $T$ on $F$ such that $T \supseteq H \cup P$.*

**Definition 6** (Corresponding Subscription). *Let $\langle \mathcal{F}_s, \mathcal{H}_s, \mathcal{F}_t, \mathcal{H}_t \rangle$ be an original catalogue and $\langle \mathcal{F}_c, \mathcal{H}_c \rangle \equiv \langle \mathcal{F}_s \cup \mathcal{F}_t, \mathcal{H}_s \cup \widehat{\mathcal{H}_t} \rangle$ be its reformulation. Given a feature subscription $S^o = \langle \langle F_s^o, H_s^o, P_s^o \rangle, \langle F_t^o, H_t^o, P_t^o \rangle \rangle$ of catalogue $\langle \mathcal{F}_s, \mathcal{H}_s, \mathcal{F}_t, \mathcal{H}_t \rangle$ and a feature subscription $S^r = \langle F^r, H^r, P^r \rangle$ of the catalogue $\langle \mathcal{F}_c, \mathcal{H}_c \rangle$, we say that $S^r$ corresponds to $S^o$ if the following holds: $F^r = F_s^o \cup F_t^o$, $H^r = H_s^o \cup \widehat{H_t^o}$, and $P^r = P_s^o \cup \widehat{P_t^o}$.*

Due to the composition of the source and target catalogues into a single catalogue, a feature subscription is consistent if and only if both source and target regions are consistent in the DFC sense.

**Proposition 1** (Equivalence of Subscription Consistency). *Let $\langle \mathcal{F}_s, \mathcal{H}_s, \mathcal{F}_t, \mathcal{H}_t \rangle$ be an original catalogue and $\langle \mathcal{F}_c, \mathcal{H}_c \rangle \equiv \langle \mathcal{F}_s \cup \mathcal{F}_t, \mathcal{H}_s \cup \widehat{\mathcal{H}_t} \rangle$ be its reformulation. A feature subscription $S^o = \langle \langle F_s^o, H_s^o, P_s^o \rangle, \langle F_t^o, H_t^o, P_t^o \rangle \rangle$ of catalogue $\langle \mathcal{F}_s, \mathcal{H}_s, \mathcal{F}_t, \mathcal{H}_t \rangle$ is consistent if and only if the corresponding subscription $S^r = \langle F^r, H^r, P^r \rangle$ of catalogue $\langle \mathcal{F}_c, \mathcal{H}_c \rangle$ is consistent.*

*Proof.* From Definition 6 we have $F^r = F_s^o \cup F_t^o$, $H^r = H_s^o \cup \widehat{H_t^o}$, and $P^r = P_s^o \cup \widehat{P_t^o}$.

($\Rightarrow$) If $S^r$ is consistent then there exists a total order $T^r$ on $F^r$ such that $T^r \supseteq H^r \cup P^r$. Let $T_s^o = T^r \downarrow_{F_s^o}$ and let $T_t^o = \widehat{T^r \downarrow_{F_t^o}}$. Both $T_s^o$ and $T_t^o$ are total orders on $F_s^o$ and $F_t^o$ respectively. Since, $T_s^o \supseteq H_s^o \cup P_s^o$, $T_t^o \supseteq H_t^o \cup P_t^o$, and $T_s^o$ is equivalent to $\widehat{T_t^o}$ on $F_s^o \cap F_t^o$, $S^o$ is also consistent.

($\Leftarrow$) If $S^o$ is consistent then there exist two total orders $T_s^o$ and $T_t^o$ on $F_s^o$ and $F_t^o$ respectively such that $T_s^o \supseteq H_s^o \cup P_s^o$, $T_t^o \supseteq H_t^o \cup P_t^o$, and $T_s^o$ is equivalent to $\widehat{T_t^o}$ on $F_s^o \cap F_t^o$. We will prove that $T_s^o \cup \widehat{T_t^o}$ is acyclic. This implies that $S^r$ is consistent (see Definition 5), since $T^r \supseteq H^r \cup P^r$, where $T^r$ is any total order on $F^r$ extending $T_s^o \cup \widehat{T_t^o}$. Note that, for $f, f' \in F_s^o$ we have $\langle f, f' \rangle \in T_s^o \cup \widehat{T_t^o}$ if and only if $\langle f, f' \rangle \in T_s^o$. We will prove that $T_s^o \cup \widehat{T_t^o}$ is acyclic by contradiction. Assume that $T_s^o \cup \widehat{T_t^o}$ is not acyclic. Thus there exists a cycle, and, in particular, a cycle of minimum cardinality, say, $k$. Therefore there exists some $f_1, \ldots, f_k, \in F^r$ such that for all $i = 0, \ldots, k$, $\langle f_i, f_{i+1} \rangle \in T_s^o \cup \widehat{T_t^o}$, where we define $f_{k+1} = f_1$ and $f_0 = f_k$. Suppose that $f_i \in F_s^o \setminus F_t^o$ for some $i \geq 1$. Then, we must have $\langle f_{i-1}, f_i \rangle \in T_s^o$ and $\langle f_i, f_{i+1} \rangle \in T_s^o$ which implies that $\langle f_{i-1}, f_{i+1} \rangle \in T_s^o$ by transitivity of $T_s^o$. But then we still have a cycle if we omit $f_i$, which contradicts the minimality of the cycle length $k$. We have shown, for all $i \geq 1$, that $f_i \in F_t^o$ and so $\langle f_i, f_{i+1} \rangle \in T_t^o$. Transitivity of $T_t^o$ implies that $\langle f_1, f_{k+1} \rangle \in T_t^o$, i.e., $\langle f_1, f_1 \rangle \in T_t^o$, which contradicts $T_t^o$ being a strict total order. □

**Proposition 2** (Complexity of Consistency Checking). *Determining whether a feature subscription $\langle F, H, P \rangle$ is consistent or not can be checked in $\mathcal{O}(|F| + |H| + |P|)$.*

*Proof.* We use *Topological Sort* (Cormen, Leiserson, & Rivest, 1990). In Topological Sort we are interested in ordering the nodes of a directed graph such that if a directed edge $\langle i, j \rangle$ is in the set of edges of the graph then node $i$ is less than node $j$ in the order. In order to use *Topological Sort* for detecting whether a feature subscription is consistent, we associate nodes with features and edges with precedence constraints. Then, the subscription is consistent if and only if for all edges $\langle i, j \rangle$ in the graph associated with the subscription, $i$ precedes $j$ in the order computed by *Topological Sort*. As the complexity of *Topological Sort* is linear with respect to the size of the graph (i.e., the sum of the number of nodes and the number of edges of the graph) detecting whether a feature subscription is consistent is $\mathcal{O}(|F| + |H| + |P|)$. □

**Definition 7** (Anti-subscription). *Given a catalogue $\langle \mathcal{F}_c, \mathcal{H}_c \rangle$ and a consistent feature subscription $S = \langle F, H, P \rangle$, the* anti-subscription *is the tuple $\langle F_a, P_a \rangle$ defined as follows. $f \in \mathcal{F}_c$ is an element of $F_a$ if and only if the directed graph associated with the subscription obtained after adding feature $f$, i.e., $\langle F \cup \{f\}, \mathcal{H}_c \downarrow_{F \cup \{f\}} \cup P \rangle$, is cyclic; $\forall i, j \in F$, $i \prec j$ is in $P_a$ if and only if the directed graph associated with the subscription obtained after adding precedence $i \prec j$, i.e., $\langle F \cup \{i, j\}, \mathcal{H}_c \downarrow_{F \cup \{i, j\}} \cup P \cup \{i \prec j\} \rangle$, is cyclic.*

The definition of anti-subscription suggests one way of computing the anti-subscription of a given subscription. In order to test whether a feature/precedence belongs to the anti-subscription we check the consistency of the resulting subscription. As there are $\mathcal{O}(|\mathcal{F}_c|)$ features and $\mathcal{O}(|\mathcal{F}_c|^2)$ precedences, the worst-case time complexity of computing an anti-subscription is at most $\mathcal{O}(|\mathcal{F}_c|^2 \times (|F| + |H| + |P|))$.

**Definition 8** (Partial Order of a Consistent Subscription). *Given a consistent subscription $\langle F, H, P \rangle$, the partial order of the subscription is the transitive closure $(H \cup P)^*$ of the relation $H \cup P$.*

The worst-case complexity of finding this transitive closure is $\mathcal{O}(|F|^3)$.

**Definition 9** (Total Order of a Consistent Subscription). *A total order of consistent subscription $S$ is a topological sort of the directed graph $\langle F, H \cup P \rangle$, i.e., a total order extending the relation $H \cup P$.*

The worst-case complexity of finding such a total order is linear in time with respect to the size of the corresponding graph.

## 5. Symmetry Inherent in the Reformulation

One of the services provided to an end-user when configuring a feature subscription is the computation of all compatible pairs of total orders on source and target features. In this section, we show that when an original subscription, as defined in Section 3, is reformulated, as described in Section 4, symmetries are introduced. Two total orders in the reformulated subscription are symmetric if they correspond to the same pair of total orders (on source and target features) in the original subscription. More formally, let $S^o = \langle \langle F_s^o, H_s^o, P_s^o \rangle, \langle F_t^o, H_t^o, P_t^o \rangle \rangle$ be a subscription of the catalogue $\langle \mathcal{F}_s, \mathcal{H}_s, \mathcal{F}_t, \mathcal{H}_t \rangle$, and $S^r = \langle F^r, H^r, P^r \rangle$ be the corresponding subscription of the catalogue $\langle \mathcal{F}_c, \mathcal{H}_c \rangle \equiv \langle \mathcal{F}_s \cup \mathcal{F}_t, \mathcal{H}_s \cup \widehat{\mathcal{H}_t} \rangle$, i.e., $F^r = F_s^o \cup F_t^o$, $H^r = H_s^o \cup \widehat{H_t^o}$, and $P^r = P_s^o \cup \widehat{P_t^o}$. A pair of total orders $\langle T_s, T_t \rangle$ is compatible with $S^o$ if Conditions (1), (2) and (3) of Definition 3 hold. There is a many-to-one relation between the set of total orders of $S^r$ (see Definition 9) and the set of compatible pairs of total orders of $S^o$.

Let us consider the subscription $S^o$ where $F_s^o = \{1, 2, 3\}$, $F_t^o = \{2, 3, 4\}$, $H_s^o = \{1 \prec 2\}$, $H_t^o = \{4 \prec 3\}$, $P_s^o$ and $P_t^o$ are empty. The corresponding $S^r$ would have $F^r = \{1, 2, 3, 4\}$, $H^r = \{1 \prec 2, 3 \prec 4\}$, and $P^r = \emptyset$. Both $S^o$ and $S^r$ are consistent. The set of total orders of $S^r$, and the set of compatible pairs of total orders of $S^o$ are shown in Table 1. The cardinality of the former set is six, while for the latter is only five. The last two total orders of $S^r$ correspond to the last compatible pair of total orders of $S^o$. This is due to the fact that the union of a total order on source features and the transpose of a total order on target features in $S^o$ is not necessarily a total order. For example for the last pair of total orders of $S^o$ in Table 1, the union of $3 \prec 1 \prec 2$ and $3 \prec 4 \prec 2$ do not result in a total order, since there is no order between 1 and 4.

The repetition of the computation of the symmetric pairs of total orders of the original subscription from the total orders of the reformulated subscription is not desirable. In order to compute a compatible pair of total orders only once, we use the algorithm GetSolutions($S^r$), as shown in Algorithm 1. This algorithm has two nested loops. In the first loop it selects a total order on the set of reversible features and then extends this total order to generate a set of total orders on source features and a set of total orders on target features. In the second loop a total order on source features and a total order on target features are selected from the previously generated sets. Due to the fact that the source features and the target features are ordered independently in GetSolutions($S^r$), no unnecessary ordering is imposed between the source features and the target features.

Table 1: Total orders on $F^r$, $F_s^o$, and $F_t^o$.

| $S^r$ | $S^o$ | |
|---|---|---|
| $F^r$ | $F_s^o$ | $F_t^o$ |
| $1 \prec 2 \prec 3 \prec 4$ | $1 \prec 2 \prec 3$ | $4 \prec 3 \prec 2$ |
| $1 \prec 3 \prec 2 \prec 4$ | $1 \prec 3 \prec 2$ | $4 \prec 2 \prec 3$ |
| $1 \prec 3 \prec 4 \prec 2$ | $1 \prec 3 \prec 2$ | $2 \prec 4 \prec 3$ |
| $3 \prec 1 \prec 2 \prec 4$ | $3 \prec 1 \prec 2$ | $4 \prec 2 \prec 3$ |
| $3 \prec 1 \prec 4 \prec 2$ | $3 \prec 1 \prec 2$ | $2 \prec 4 \prec 3$ |
| $3 \prec 4 \prec 1 \prec 2$ | | |

---

**Algorithm 1** GetSolutions($S^r$)

**Require:**

- $S^r = \langle F^r, H^r, P^r \rangle$ is a consistent subscription, where $F^r = F_s^o \cup F_t^o$, $F_s^o$ is the set of source features, $F_t^o$ is the set of target features, and $F_r^o = F_s^o \cap F_t^o$ is the set of reversible features in a corresponding subscription $S_o$.

- GetTotalOrders($\langle F, O \rangle$) generates the set of all total orders that extend a given acyclic binary relation $O$ defined on a set of features $F$.

- $\succ$, $\succ_R$, $\succ_S$, and $\succ_T$ are set to $(H^r \cup P^r)^*$, $\succ\downarrow_{F_r^o}$, $\succ\downarrow_{F_s^o}$, and $\succ\downarrow_{F_t^o}$ respectively.

**Ensure:** $PTOs$ is the set of pairs of compatible total orders on $F_s^o$ and $F_t^o$ respectively.

1: $PTOs \leftarrow \emptyset$
2: $RTOs \leftarrow$ GetTotalOrders($\langle F_r^o, \succ_R \rangle$)
3: **for all** $\succ_r \in RTOs$ **do**
4:     $STOs \leftarrow$ GetTotalOrders($\langle F_s^o, \succ_S \cup \succ_r \rangle$)
5:     $TTOs \leftarrow$ GetTotalOrders($\langle F_t^o, \succ_T \cup \succ_r \rangle$)
6:     **for all** $\succ_s \in STOs, \succ_t \in TTOs$ **do**
7:         $PTOs \leftarrow PTOs \cup \{\langle \succ_s, \widehat{\succ_t} \rangle\}$
8: **return** $PTOs$

---

The algorithm computes and saves all total orders on a given set of reversible features in $RTOs$, and for a given total order on the set of reversible features it computes and saves all the total orders on source and target features in $STOs$ and $TTOs$ respectively. However, this is presented in the algorithm for the purpose of clarity. In practice, a total order is computed lazily, i.e., a total order is only computed when is needed, thus avoiding the need of keeping all the total orders generated in memory.

The amortised time complexity of computing all the total orders extending a given acyclic binary relation is linear with respect to the number of total orders (Pruesse & Ruskey, 1994). Assuming that there are $\tau_r$ total orders on $F_r^o$ and at most $\tau_s$, and $\tau_t$ total orders on $F_s^o$ and $F_t^o$ that are consistent with a given total order on $F_r^o$ respectively, the time complexity of GETSOLUTIONS is $\mathcal{O}(\tau_r \times \tau_s \times \tau_t)$. The computation of all the pairs of compatible total orders could be impractical when the size of the resulting set is very large. Therefore, in those cases the computation of the number of total orders could be restricted to a pre-specified number, and a heuristic can be used to select $\succ_r$ in Line 3, and $\succ_s$ in Line 6 of Algorithm 1.

There may be some pairs of total orders on $F_s^o$ and $F_t^o$ that are more desirable than others. For instance, it would be more desirable to present an end-user those pairs of total orders that are more easy to extend (in terms of the addition of a feature or a user precedence). One way of doing this is to use the notion of *anti-subscription* (see Definition 7). Each pair of total orders can be associated with an anti-subscription. The size of the anti-subscription is the sum of the number of features and precedences that are involved in it. The pairs of total orders can be ordered in the increasing size of their corresponding anti-subscriptions. The size of an anti-subscription in some sense reflects how constrained a pair of total orders is with respect to the future addition of the number of features and user precedences that an end-user may consider in his/her subscription in the future.

## 6. Relaxations of Feature Subscriptions

If an input feature subscription is not consistent then the goal is to relax it by dropping one or more features or user precedence constraints to generate a consistent feature subscription that is closest to the initial user's requirements. Therefore, we introduce a function $w : F \cup P \to \mathbb{N}$ that assigns weights to features and user precedence constraints, indicating the importance to the user of the features and user precedences. These weights could be elicited directly through data mining or analysis of user interactions. In the rest of the paper a feature subscription is denoted by $S = \langle F, H, P, w \rangle$. The value of the subscription $S$ is defined by $\text{Value}(S) = \sum_{f \in F} w(f) + \sum_{\rho \in P} w(\rho)$.

**Definition 10** (Relaxation). *A relaxation of a feature subscription $\langle F, H, P, w \rangle$ of a catalogue $\langle \mathcal{F}_c, \mathcal{H}_c \rangle$ is a subscription $\langle F', H', P', w' \rangle$ such that $F' \subseteq F$, $H' = H \downarrow_{F'}$, $P' \subseteq P \downarrow_{F'}$ and $w'$ is $w$ restricted to $F' \cup P'$.*

**Definition 11** (Optimal Relaxation). *Let $R_S$ be the set of all consistent relaxations of a feature subscription $S$. We say that $S_i \in R_S$ is an optimal relaxation of $S$ if it has maximum value among all consistent relaxations, i.e., if and only if there does not exist $S_j \in R_S$ such that $\text{Value}(S_j) > \text{Value}(S_i)$.*

**Proposition 3** (Complexity of Finding an Optimal Relaxation). *Finding an optimal relaxation of a feature subscription is NP-hard.*

*Proof.* Given a directed graph $G = \langle V, E \rangle$, the *Feedback Vertex Set Problem* is to find a smallest $V' \subseteq V$ whose deletion makes the graph acyclic. This problem is known to be NP-hard (Garey & Johnson, 1979). We prove that finding an optimal relaxation is NP-hard by a reduction from the feedback vertex set problem. The feedback vertex set problem can be reduced to our problem by associating the nodes of the directed graph $V$ with features $F$, the edges $E$ with catalogue precedence constraints $H$. We set $P$ to $\emptyset$ and define $w$ by $w(f) = 1$, for all $f \in F$. Thus, finding an optimal relaxation of $S = \langle F, H, P, w \rangle$ corresponds to finding a biggest set of nodes $V''$ such that the deletion of $V - V''$ from $G$ results in an acyclic graph. Therefore, we conclude that finding an optimal relaxation of an inconsistent subscription is NP-hard. □

The most challenging operation on feature subscriptions is to find an optimal relaxation of a subscription that is not consistent, since it is NP-Hard. In the remainder of the paper we focus only on this particular task.

## 7. Basic COP Model for Finding an Optimal Relaxation

In this section we model the problem of finding an optimal relaxation of a feature subscription $\langle F, H, P, w \rangle$ of catalogue $\langle \mathcal{F}_c, \mathcal{H}_c \rangle$ as a constraint optimisation problem (Lesaint, Mehta, O'Sullivan, Quesada, & Wilson, 2008c).

**Variables and Domains.** We associate each feature $i \in F$ with two variables: a *Boolean variable* $bf_i$ and an *integer variable* $pf_i$. A Boolean variable $bf_i$ is instantiated to 1 or 0 depending on whether feature $i$ is included in the subscription or not, respectively. The domain of each integer variable $pf_i$ is $\{1, \ldots, |F|\}$. Assuming that the computed subscription is consistent, an integer variable $pf_i$ corresponds to the position of the feature $i$ in a sequence, which is consistent with the optimal relaxation. We associate each user precedence constraint $(i \prec j) \in P$ with a *Boolean variable* $bp_{ij}$. A Boolean variable $bp_{ij}$ is instantiated to 1 or 0 depending on whether $i \prec j$ is respected in the computed subscription or not, respectively. A variable $v$ is associated with the value of the subscription, the initial lower bound of which is 0 and the initial upper bound is the sum of the weights of all the features and user precedences.

**Constraints.** A catalogue precedence constraint $(i \prec j) \in H$ that feature $i$ should be before feature $j$ can be expressed as follows:

$$bf_i \wedge bf_j \;\Rightarrow\; (pf_i < pf_j).$$

Note that the constraint is activated only if the selection variables $bf_i$ and $bf_j$ are instantiated to 1. A user precedence constraint $(i \prec j) \in P$ that $i$ should be placed before $j$ in their subscription can be expressed as follows:

$$bp_{ij} \Leftrightarrow (bf_i \wedge bf_j \wedge (pf_i < pf_j)).$$

Note that if a user precedence constraint holds then the features $i$ and $j$ are included in the subscription and also the feature $i$ is placed before $j$, that is, the selection variables $bf_i$ and $bf_j$ are instantiated to 1 and $pf_i < pf_j$ is true.

The value of the subscription is equal to the sum of the weights of the included features and included user precedences. This constraint can be expressed as the following:

$$v = \sum_{i \in F} bf_i \times w(i) + \sum_{(i \prec j) \in P} bp_{ij} \times w(i \prec j). \tag{1}$$

Enforcing arc consistency on Equation (1), in general, is exponential (Zhang & Yap, 2000). Therefore, CP solvers perform only bounds consistency on this constraint, which is equivalent to enforcing arc consistency on the the following pair of constraints, which can be seen as a decomposition of Equation (1):

$$v \geq \sum_{i \in F} bf_i \times w(i) + \sum_{(i \prec j) \in P} bp_{ij} \times w(i \prec j). \tag{2}$$

$$v \leq \sum_{i \in F} bf_i \times w(i) + \sum_{(i \prec j) \in P} bp_{ij} \times w(i \prec j). \tag{3}$$

In order to reason about the complexities of enforcing different consistency techniques we always assume that the two inequality constraints are used instead of the equality constraint.

**Objective.** The objective is to find an optimal relaxation of a feature subscription.

We have investigated the impact of maintaining three different levels of consistency within branch and bound search. The first is arc consistency and the rest are mixed consistencies. In the following sections we shall describe these consistency techniques and present their worst-case time complexities when enforced on any instance of feature subscription, if formulated as described above. The results for the complexities that are presented below are based on the assumption that only the Boolean variables associated with the inclusion/exclusion of features and user precedences are the decision variables. We remark that if the problem is arc-consistent after instantiating all the Boolean variables then it is also globally consistent.

## 7.1 Arc Consistency

Let $e$ be the sum of the number of user precedences and the number of catalogue precedences, let $n$ be the sum of the number of features and the number of user precedences, and let $d$ be the number of features. The complexity of achieving arc consistency (AC) on a (catalogue/user) precedence constraint is constant with respect to the number of variables. A catalogue precedence constraint is made arc-consistent when any of the Boolean variables involved in the constraint is initialised or any of the domains of the position variables is modified. Thus, a catalogue precedence constraint can be made arc-consistent at most $(1 + 1 + (d-1) + (d-1))$ times, which is effectively $2d$ times. A user precedence constraint can be made arc-consistent at most $2d + 1$ times. Since there are, in total, $e$ precedence constraints, the worst-case time complexity of imposing arc consistency on all the precedence constraints is $\mathcal{O}(e\,d)$, which is also optimal. In addition, arc consistency is also enforced on the linear inequalities (2) and (3), the complexity of which is linear with respect to the number of Boolean variables. Whenever a Boolean variable is instantiated the constraint is revised and since there are $n$ Boolean variables, it can be made arc-consistent at most $n$ times. Therefore, the worst-case time complexity of enforcing arc consistency on the linear inequalities is $\mathcal{O}(n^2)$, which is optimal. Thus, the worst-case time complexity of enforcing AC on an instance of basic CP model for finding an optimal relaxation is $\mathcal{O}(e\,d + n^2)$.

## 7.2 Singleton Arc Consistency

Maintaining a higher level of consistency can be expensive in terms of time. However, if more values can be removed from the domains of the variables, the search effort can be reduced and this may save time. We shall investigate the effect of maintaining Singleton Arc Consistency (SAC) on the Boolean variables and AC on the remaining variables and denote it by $SAC_b$. We have used the SAC-1 (Debruyne & Bessiere, 1997) algorithm for enforcing SAC on the Boolean variables. Enforcing SAC on the Boolean variables in a SAC-1 manner works by traversing a list of $2n$ variable-value pairs. For each instantiation of a Boolean variable $x$ to each value $0/1$, if there is a domain wipeout while enforcing AC then the value is removed from the corresponding domain and AC is enforced. Each time a value is removed, the list is traversed again. Since there are $2n$ variable-value pairs, the number of calls to the underlying arc consistency algorithm is at most $4n^2$. Thus the worst-case time complexity of $SAC_b$ is $\mathcal{O}(n^2\,(e\,d+n^2))$.

$SAC_b$ does not have an optimal worst-case time complexity. In $SAC_b$ arc consistency can be enforced on a subproblem obtained by restricting a Boolean variable to a single value at most $2n$ times, and each time arc consistency is established from scratch. However, one can take the incremental property of arc consistency into account to obtain an optimal version of $SAC_b$. Following the work of Lecoutre (2009) an arc consistency algorithm is said to be incremental if and only if its worst-case time complexity is the same when it is applied once on a given network $P$ and when it is applied up to $m$ times on $P$ where between any two consecutive executions, at least one value has been deleted. Here $m$ is the sum of the domain sizes of all the variables involved in the problem $P$. The idea behind an optimal version is that we do not want to achieve arc consistency from scratch in each subproblem, but, instead, benefit from the incremental property of the underlying arc consistency algorithm. This results in the asymptotic complexity of $\mathcal{O}(e\,d+n^2)$ for enforcing arc consistency $2n$ times. Thus, the time complexity of an optimal version of $SAC_b$ would be $\mathcal{O}(n\,(e\,d+n^2))$.

### 7.3 Restricted Singleton Arc Consistency

The main problem with SAC-1 is that deleting a single value triggers the loop again. The Restricted Singleton Arc Consistency (RSAC) avoids this by considering each variable-value pair only once (Prosser, Stergiou, & Walsh, 2000). We investigate the effect of enforcing (RSAC) on the Boolean variables and AC on the remaining variables, and denote it by $RSAC_b$. The worst-case time complexity of $RSAC_b$ is $\mathcal{O}(n\,(e\,d+n^2))$.

## 8. Other CP Models

In this section we present two more CP approaches. The first approach uses a global constraint that achieves a higher level of consistency by taking into account the cycles of the precedence constraints. In the second approach we model the problem as a weighted constraint satisfaction problem.

### 8.1 Global Constraint

A global constraint captures a relation between several variables. It takes into account the structure of the problem to prune more values. For instance, if a user has selected a set of features, $F = \{1, 2, 3, 4\}$ and if these features are constrained by the catalogue precedences $1 \prec 2$, $2 \prec 1$, $3 \prec 4$ and $4 \prec 3$, and if three features are required to be included in the subscription then one can infer that the problem is inconsistent without doing any search.

This is possible by inferring cycles from the precedence constraints and using them to prune the bounds of the objective function.

The soft global precedence constraint SoftPrec was proposed by Lesaint et al. (2008a). It holds if and only if there is a strict partial order on the selected features subject to the relevant hard (catalogue) precedence constraints and the selected soft (user) precedence constraints, and the value of the subscription is within the provided bounds. As shown by Lesaint et al. (2008a), achieving AC for SoftPrec is NP-complete since there is no way to determine in polynomial time whether there is a strict partial order whose value is between the given bounds. Therefore, AC is approximated by pruning the domains of the variables based on the filtering rules that follow from the definition of SoftPrec. The time-complexity for achieving this pruning is $\mathcal{O}(|F|^3)$, which is polynomial. The upper bound of the value of the subscription is pruned based on the incompatibilities that are inferred between pairs of features, and the dependencies between user precedences and their corresponding features. The pruning rules of SoftPrec are used within branch and bound search to find an optimal relaxation of a feature subscription.

Let $\langle F, H, P, w \rangle$ be a subscription. Let $bf$ be a vector of Boolean variables associated with $F$. We say that feature $i$ is included if $bf(i) = 1$, and $i$ is excluded if $bf(i) = 0$. We abuse the notation by using $bf(i)$ to mean $bf(i) = 1$, and $\neg bf(i)$ to mean $bf(i) = 0$. A similar convention is adopted for the other Boolean variables. Let $bp$ be a $|F|^2$ matrix of Boolean variables. Here $bp$ is intended to represent a strict partial order on the included features $F'$ which is compatible with the catalogue constraints restricted to $F'$.

**Definition 12** (SoftPrec). *Let $S = \langle F, H, P, w \rangle$ be a feature subscription, $bf$ and $bp$ be vectors of Boolean variables, and $v$ be an integer variable, SoftPrec$(S, bf, bp, v)$ holds if and only if*

1. *$bp$ is a strict partial order restricted to $bf$, i.e.,*

$$\begin{aligned} &\forall i, j \in F : bp(i, j) \Rightarrow bf(i) \wedge bf(j) &&\text{(restricted)}, \\ &\forall i, j \in F : bp(i, j) \Rightarrow \neg bp(j, i) &&\text{(asymmetric)}, \\ &\forall i, j, k \in F : bp(i, j) \wedge bp(j, k) \Rightarrow bp(i, k) &&\text{(transitive)}, \end{aligned}$$

2. *$bp$ is compatible with $H$ restricted to $bf$, i.e.,*

$$\forall (i \prec j) \in H : bf(i) \wedge bf(j) \Rightarrow bp(i, j),$$

3. *$v = \sum_{i \in F} bf(i) \times w(i) + \sum_{(i \prec j) \in P} bp(i, j) \times w(i \prec j)$.*

The set of constraints in this CP model only contains SoftPrec. The decision variables in this model are $bf$ and $bp$. A solution of SoftPrec is a consistent relaxation of the subscription $\langle F, H, P, w \rangle$. Notice that the feedback vertex set problem (Garey & Johnson, 1979) can be expressed in terms of SoftPrec by associating vertices with features and arcs with catalogue precedence constraints. Therefore, achieving generalised arc consistency on SoftPrec is NP-hard.

## 8.2 Weighted CSP Model

The classical CSP framework has been extended by associating weights (or costs) with tuples (Larrosa, 2002). The Weighted Constraint Satisfaction Problem (WCSP) is a specific extension that relies on a specific valuation structure $S(k)$ defined as follows.

**Definition 13** (Valuation Structure). $S(k)$ *is a triple* $(\{0, \ldots, k\}, \oplus, \geq)$ *where:* $k \in \{1, \ldots, \infty\}$ *is either a strictly positive natural number or infinity,* $\{0, 1, \ldots, k\}$ *is the set of naturals less than or equal to* $k$, $\oplus$ *is the sum over the valuation structure defined as:* $a \oplus b = \min\{k, a+b\}$, $\geq$ *is the standard order among naturals.*

A WCSP instance is defined by a valuation structure $S(k)$, a set of variables (as for classical CSP instances) and a set of constraints. A domain is associated with each variable and a cost function with each constraint. More precisely, for each constraint $C$ and each tuple $t$ that can be built from the domains associated with the variables involved in $C$, a value in $\{0, 1, \ldots, k\}$ is assigned to $t$. When a constraint $C$ assigns the cost $k$ to a tuple $t$, it means that $C$ forbids $t$. Otherwise, it is permitted by $C$ with the corresponding cost. The cost of an instantiation of variables is the sum (using operator $\oplus$) over all constraints involving variables instantiated. An instantiation is consistent if its cost is strictly less than $k$. The goal of the WCSP problem is to find a full consistent assignment of variables with minimum cost. A WCSP formulation for finding an optimal relaxation of the input subscription $\langle F, H, P, w \rangle$, when inconsistent, is outlined below.

The maximum acceptable cost is

$$k = \sum_{i \in F} w(i) + \sum_{\rho \in P} w(\rho).$$

We associate each feature $i \in F$ with an *integer variable* $pf_i$. The domain of each integer variable, $D(pf_i)$, is $\{0, \ldots, |F|\}$. If $pf_i$ is instantiated to 0, it indicates that $i$ is excluded from the subscription.

A unary cost function $C_i : D(pf_i) \rightarrow \{0, w(i)\}$ assigns costs to assignments of variable $pf_i$ in the following way:

$$C_i(a) = \begin{cases} 0 & \text{if } a > 0 \\ w(i) & \text{if } a = 0 \end{cases}$$

A catalogue precedence constraint $(i \prec j) \in H$ is associated with a binary cost function $H_{i \prec j} : D(pf_i) \times D(pf_j) \rightarrow \{0, k\}$ that assigns costs to assignments of variables $pf_i$ and $pf_j$ in the following way:

$$H_{i \prec j}(a, b) = \begin{cases} 0 & \text{if } a = 0 \vee b = 0 \vee a < b \\ k & \text{otherwise} \end{cases}$$

A user precedence constraint $(i \prec j) \in P$ is associated with a binary cost function $P_{i \prec j} : D(pf_i) \times D(pf_j) \rightarrow \{0, w(i \prec j)\}$ assigns costs to assignments of variables $pf_i$ and $pf_j$ in the following way:

$$P_{i \prec j}(a, b) = \begin{cases} 0 & \text{if } a \neq 0 \wedge b \neq 0 \wedge a < b \\ w(i \prec j) & \text{otherwise} \end{cases}$$

Note that if a user precedence constraint holds then the features $i$ and $j$ are included in the subscription and also the feature $i$ is placed before $j$, that is, the integer variables $pf_i$ and $pf_j$ are instantiated to any value greater than 0 and $pf_i < pf_j$ is true.

## 9. Boolean Satisfiability

The Boolean Satisfiability Problem (SAT) is a decision problem an instance of which is an expression in propositional logic. The problem is to decide whether there is an assignment of *true* and *false* values to the variables that will make the expression *true*. The expression is normally written in conjunctive normal form. The Partial Weighted Maximum Boolean Satisfiability Problem (PWMSAT) is an extension of SAT that includes the notions of hard and soft clauses. Any solution should respect the hard clauses. Soft clauses are associated with weights. The goal is to find an assignment that satisfies all the hard clauses and minimises the sum of the weights of the unsatisfied soft clauses. In this section we present Boolean satisfiability formulations for finding an optimal relaxation of a feature subscription.

### 9.1 Atom-based Encoding

In an atom-based encoding, each atom, like $f \prec g$, is associated with a propositional variable and the asymmetricity and transitivity properties of the precedence relation are explicitly encoded. An atom-based encoding of finding an optimal relaxation of a feature subscription $\langle F, H, P, w \rangle$ is outlined below.

**Variables.** Let *PrecDom* be the set of possible precedence constraints that can be defined on $F$, i.e., $\{i \prec j : \{i, j\} \subseteq F \land i \neq j\}$. For each feature $i \in F$ there is a Boolean variable $bf_i$, which is true or false depending on whether feature $i$ is included or not in the computed subscription. For each precedence constraint $(i \prec j)$ there is a Boolean variable $bp_{ij}$, which is true or false depending on whether the precedence constraint holds or not in the computed subscription. If $bp_{ij}$ is true, then, roughly speaking, it means that features $i$ and $j$ are included, and $i$ precedes $j$.

**Clauses.** Each weighted-clause is represented by a tuple $\langle w, c \rangle$, where $w$ is the weight of the clause $c$. Note that the hard clauses are associated with weight $\top$, which represents an infinite penalty for not satisfying them.

Each catalogue precedence constraint, $(i \prec j) \in H$, must be satisfied if the features $i$ and $j$ are included in the computed subscription. This is modelled by adding the following hard clause:

$$\langle \top, (\neg bf_i \lor \neg bf_j \lor bp_{ij}) \rangle.$$

The precedence relation should be transitive and asymmetric in order to ensure that the subscription graph is acyclic. To ensure asymmetricity, the following clause is added for every pair $\{i \prec j, j \prec i\} \subseteq PrecDom$:

$$\langle \top, (\neg bp_{ij} \lor \neg bp_{ji}) \rangle. \tag{4}$$

Both $bp_{ij}$ and $bp_{ji}$ can be false. However, if one of them is true the other one should be false.

To ensure transitivity, for every $\{i \prec j, j \prec k\} \subseteq PrecDom$, the following clause is added:

$$\langle \top, (\neg bp_{ij} \lor \neg bp_{jk} \lor bp_{ik}) \rangle. \tag{5}$$

Note that Rule (5) need only be applied to $\langle i, j, k \rangle$ such that $i \neq k$ since precedence constraints are not reflexive because of Rule (4).

Each precedence constraint $(i \prec j) \in PrecDom$ is only satisfied when its corresponding features $i$ and $j$ features are included. This is ensured by considering the following clauses:

$$\langle \top, (\neg bp_{ij} \vee bf_i) \rangle \quad \langle \top, (\neg bp_{ij} \vee bf_j) \rangle.$$

We need to penalise any solution that does not include a feature $i \in F$ or a user precedence constraint $(i \prec j) \in P$. This is done by adding the following clauses:

$$\langle w(i), (bf_i) \rangle \quad \langle w(i \prec j), (bp_{ij}) \rangle.$$

The cost of violating these clauses is the weight of the feature $i$ and the weight of the user precedence constraint $i \prec j$ respectively.

**Reducing the Variables and Clauses.** It is straightforward to realise that the atom based encoding described in the previous section requires $\Theta(n^2)$ Boolean variables and $\Theta(n^3)$ clauses, where $n$ is the number of features[1]. We now describe two techniques which can reduce the number of variables and clauses. The subscription contains a cycle if and only if the transitive closure of $H \cup P$ contains a cycle. Therefore, instead of associating a Boolean variable with each possible precedence constraint, it is sufficient to associate Boolean variables only with the precedence constraints in the transitive closure of $H \cup P$. Reducing the Boolean variables will also reduce the transitive clauses, especially when the input subscription graph is not dense. Otherwise, Rule (5) will generate $|F| \times (|F| - 1) \times (|F| - 2)$ transitivity clauses and Rule (4) will generate $(|F| \times (|F| - 1))/2$ asymmetricity clauses. For example, for the subscription $\langle F, H, P, w \rangle$ with $F = \{1, 2, 3, 4, 5, 6\}$, $H = \{1 \prec 2, 2 \prec 1, 3 \prec 4, 4 \prec 3, 5 \prec 6, 6 \prec 5\}$, and $P = \emptyset$, Rules (4) and (5) will generate 120 transitivity clauses and 15 asymmetricity clauses respectively. Since any relaxation of the given subscription respecting the clauses generated by Rule (4) is acyclic, the 120 transitivity clauses and 12 asymmetricity clauses are redundant. Thus, if $PrecDom$ is instead set to be the transitive closure of $H \cup P$, then Rules (4) and (5) would not generate any redundant clauses. We further reduce the number of transitivity clauses $\langle \top, (\neg bp_{ij} \vee \neg bp_{jk} \vee bp_{ik}) \rangle$ by considering only those where none of $j \prec i$, $k \prec j$, and $i \prec k$ are in $H$, especially when the input subscription graph is not sparse. The reason for this is that these transitivity clauses are always entailed due to the enforcement of the catalogue precedence constraints. This reduction in the number of clauses might reduce the memory requirement and also might have an impact on the efficiency of unit propagation, which in turn may reduce the runtime.

## 9.2 Symbol-based Encoding

Another SAT approach based on a symbol-based encoding of partial order constraints is presented by Codish et al. (2009). Partial order constraints (Codish, Lagoon, & Stuckey, 2008) are basically propositional formulae except that propositions can also be statements about a partial order on a finite set of symbols. In a symbol-based encoding the transitivity and asymmetricity properties of a precedence relation are enforced implicitly.

Here also a Boolean variable $bf_i$ is associated with each feature $i \in F$ indicating whether $i$ is included or excluded. A Boolean variable $bp_{ij}$ is associated with each precedence constraint $(i \prec j) \in H \cup P$. For each catalogue constraint $(i \prec j) \in H$ the following clause is added: $\langle \top, (\neg bf_i \vee \neg bf_j \vee bp_{ij}) \rangle$. For each precedence constraint $i \prec j \in (H \cup P)$ the following clauses are added: $\langle \top, (\neg bp_{ij} \vee bf_i) \rangle$ and $\langle \top, (\neg bp_{ij} \vee bf_j) \rangle$. For each precedence constraint $i \prec j \in (H \cup P)$ the propositional constraint $bp_{ij} \Rightarrow [\![i \prec j]\!]$ is encoded[2]. This intuitively means that if $bp_{ij}$ is true then $i$ precedes $j$. Two different ways of encoding a precedence constraint $[\![i \prec j]\!]$ are presented by Codish et al. (2009), which are called the *unary encoding* and the *binary encoding*. A brief description of them is presented in Section 9.2.1 and Section 9.2.2, which will provide a basis for their theoretical comparisons.

---

1. Given a function $g(n)$, $\Theta(g(n))$ denotes the set of functions $f(n)$ such that there exist positive constants $c_1$, $c_2$ and $n_0$ such that $0 \leq c_1 g(n) \leq f(n) \leq c_2 g(n)$ for all $n \geq n_0$ (Cormen et al., 1990).

Advanced techniques for encoding the objective function have also been proposed by Codish et al. (2009). However the encoding of the objective function is orthogonal to the way the precedences are encoded. As our purpose is to compare the encoding of the precedence constraints, we omit the details of the encoding of the objective function for the symbol-based encoding proposed by Codish et al. (2009). Instead, we assume that in this approach the objective function is encoded as it is done in the atom-based case. Therefore, in the PWMSAT setting the following soft clauses are added for features and user precedences: $\langle w(i), bf_i \rangle$ and $\langle w(i \prec j), bp_{ij} \rangle$.

### 9.2.1 UNARY ENCODING

In the symbol-based unary encoding (Codish et al., 2009) each feature is associated with an ordered set of Boolean variables that represents the unary encoding of its position. The unary encoding of a non-negative integer $m \leq n$ is an assignment of values to a sequence of $n$ Boolean variables $\langle m_1, \ldots, m_n \rangle$ such that $m_1 \geq m_2 \geq \cdots \geq m_n$. The integer-value of such a representation is the number of variables $m_i$ taking value 1. For example, the sequence 11100000 represents the number $m = 3$ using $n = 8$ variables. For each pair of consecutive variables in the sequence, say $m_k$ and $m_{k+1}$, a clause $\langle \top, (\neg m_{k+1} \vee m_k) \rangle$ is introduced to the encoding in order to enforce that if $m_{k+1}$ is assigned 1 then its predecessor in the sequence, $m_k$, must be assigned 1. Let $i$ and $j$ be two non-negative integer variables that can be assigned values less than or equal to $n$. Let $\langle i_1, \ldots, i_n \rangle$ and $\langle j_1, \ldots, j_n \rangle$ be the sequences of $n$ Boolean variables that represent the unary-encodings of $i$ and $j$ respectively. The unary-encoding of $i \prec j$ is denoted by $\langle i_1, \ldots, i_n \rangle \prec \langle j_1, \ldots, j_n \rangle$, which means that the number of variables assigned the values 1 in the sequence $\langle i_1, \ldots, i_n \rangle$ is less than the number of variables assigned the values 1 in the sequence $\langle j_1, \ldots, j_n \rangle$. Notice that $\langle i_1, \ldots, i_n \rangle \prec \langle j_1, \ldots, j_n \rangle$ holds if and only if $\neg i_n$ holds, $j_1$ holds, and $\langle i_1, \ldots, i_n \rangle \preceq \langle j_2, \ldots, j_n, 0 \rangle$ holds. Here $\langle j_2, \ldots, j_n, 0 \rangle$ encodes an integer between 0 and $n - 1$, which is the predecessor of $\langle j_1, \ldots, j_n \rangle$. The inequality $\langle i_1, \ldots, i_n \rangle \preceq \langle j_2, \ldots, j_n, 0 \rangle$ can be encoded as follows: $\forall 1 \leq k \leq n-1$, $i_k \Rightarrow j_{k+1}$. The resulting weighted clauses for $bp_{ij} \Rightarrow [\![i \prec j]\!]$ are $\langle \neg bp_{ij} \vee \neg i_n \rangle$, $\langle \neg bp_{ij} \vee j_1 \rangle$, and $\forall 1 \leq k \leq n-1$, $\langle \top, (\neg bp_{ij} \vee \neg i_k \vee j_{k+1}) \rangle$. Overall, the symbol-based unary encoding requires $\Theta(n^2)$ propositional variables ($n$ per feature) and involves $\Theta(k\,n)$ clauses ($n$ per precedence constraint), where $k = |H \cup P|$.

### 9.2.2 BINARY ENCODING

In the symbol-based binary encoding each feature is associated with an ordered set of Boolean variables that represents the binary log encoding of its position. The binary encod-ing of a non-negative integer $a \leq n$ is a sequence of values assigned to $k$ variables $v_1, \ldots, v_k$, where $k = \lceil \log_2 n \rceil$. The value of such a representation is $\sum_{1 \leq m \leq k} 2^{k-m} \times v_m$. For example, the sequence 101 represents the number 5 using 3 variables. A precedence constraint is encoded using a lexicographical comparator (Apt, 2003). Given two numbers in binary encoded form $\langle i_1, \ldots, i_k \rangle$ and $\langle j_1, \ldots, j_k \rangle$, a precedence constraint $\langle i_1, \ldots, i_k \rangle < \langle j_1, \ldots, j_k \rangle$ holds if and only if there exists $m > 0$ such that $i_m < j_m$ and for all $l < m$, $i_l = j_l$. The resulting encoding is not in conjunctive normal form. Therefore, the Tseitin transformation[3] (Tseitin, 1968) is used to obtain the corresponding formula in conjunctive normal form. For a given precedence constraint, the Tseitin transformation introduces $\Theta(\log n)$ variables and clauses, since $\log n$ is the length of the formula associated with the given precedence constraint. Overall, the symbol-based binary encoding requires $\Theta(n \log n)$ propositional variables and involves $\Theta(k \log n)$ clauses, where $k = |H \cup P|$.

---

2. $[\![i \prec j]\!]$ is a Boolean formula that is satisfiable if and only if $i$ precedes $j$.

## 9.3 Comparison of the Encodings

Unit Propagation (UP) is a central component of a search-based SAT solver. Given a unit clause $l$, unit propagation applies the following rules: (1) every clause containing $l$ is removed, and (2) $\neg l$ is removed from every clause that contains this literal. These rules are applied until a fixed-point is reached. The application of these two rules leads to a new set of clauses that is equivalent to the old one. Unit propagation detects inconsistency when an empty clause is generated.

Let AE, SE$_u$, and SE$_b$ denote the atom-based encoding, the symbol-based unary encoding, and the symbol-based binary encoding respectively. The difference between these encodings is the way they encode acyclicity. In AE acyclicity is encoded explicitly by adding transitivity and asymmetricity clauses. In SE$_u$ and SE$_b$ acyclicity is encoded implicitly by associating each feature with a set of Boolean variables that represent its position (an integer value) and a precedence constraint is expressed in terms of these positions. The Boolean variables denoting the inclusion (or exclusion) of features and user precedences are called *problem variables*. These variables are common to all the encodings. An optimal relaxation can be expressed in terms of the problem variables. In order to show that unit propagation on one encoding is stronger than unit propagation on another encoding, we need to map the decisions of one encoding to the other one. Unfortunately, it is not possible to map the decisions between the atom-based and the symbol-based encodings. For example, an assignment of a position variable in the symbol-based encodings cannot be expressed in terms of the assignments to the variables of AE. Nevertheless, in the following, we prove that unit propagation in AE is stronger than unit propagation in SE$_b$ when a set of assignments are restricted to the problem variables.

**Proposition 4.** *Given a set of assignments restricted to the problem variables, if unit propagation detects inconsistency in* SE$_b$ *then it also detects inconsistency in* AE*, but the converse is not true.*

---

3. Given a propositional formula, the Tseitin transformation obtains an equivalent formula in conjunctive normal form by associating a new variable with every subformula of the original formula and applying the following equivalences: (i) $s_0 \Leftrightarrow (s_1 \vee s_2) \equiv \{(\neg s_0 \vee s_1 \vee s_2), (s_0 \vee \neg s_1), (s_0 \vee \neg s_2)\}$, (ii) $s_0 \Leftrightarrow (s_1 \wedge s_2) \equiv \{(s_0 \vee \neg s_1 \vee \neg s_2), (\neg s_0 \vee s_1), (\neg s_0 \vee s_2)\}$, and (iii) $s_0 \Leftrightarrow \neg s_1 \equiv \{(\neg s_0 \vee \neg s_1), (s_0 \vee s_1)\}$.

*Proof.* The atom-based and the symbol-based binary encoding differ only on the encoding of the acyclicity, i.e., the encoding of the transitivity and asymmetricity properties of the precedence relation. In the symbol-based binary encoding transitivity and asymmetricity properties are implicitly captured by the clauses corresponding to the propositional constraints of the form $bp_{ij} \Rightarrow [\![i \prec j]\!]$. Therefore, in order to prove that if UP detects inconsistency in SE$_b$ then it also detects inconsistency in AE, it is sufficient to show that if $bp_{ij}$ is falsified due to violation of $[\![i \prec j]\!]$ in SE$_b$ under unit propagation, the same happens in AE. The clauses corresponding to $[\![i \prec j]\!]$ are not defined in terms of the problem variables and none of these clauses are unary[4]. Therefore, UP can not falsify $bp_{ij}$ in SE$_b$. This trivially implies that, when only a set of problem variables are instantiated, UP in AE detects any inconsistency that is detected by UP in SE$_b$.

Now we show that there exists a case where an inconsistency is detected by UP in AE but it is not detected in SE$_b$. Let $F = \{i, j, k\}$ be a set of features, $H = \emptyset$, and $P = \{i \prec j, j \prec k, k \prec i\}$ be a set of user precedence constraints. In all the encodings we have a Boolean variable per user precedence constraint: $bp_{ij}$, $bp_{jk}$ and $bp_{ki}$ and we assume that $bp_{ij}$, $bp_{jk}$ and $bp_{ki}$ are set to true. In AE the unit resolution of $bp_{ij}$ and $bp_{jk}$ with the transitive clause $\neg bp_{ij} \vee \neg bp_{jk} \vee bp_{ik}$ yields $bp_{ik}$, and the unit-resolution of $bp_{ik}$ with $\neg bp_{ki} \vee \neg bp_{ik}$ yields $\neg bp_{ki}$, which results in an empty clause when resolved with $bp_{ki}$. In SE$_b$, an ordered set of Boolean variables is associated with each feature. As there are 3 features, two Boolean variables are required per feature. Therefore each feature $i$, $j$ and $k$ is associated with $\langle i_1, i_2 \rangle$, $\langle j_1, j_2 \rangle$, and $\langle k_1, k_2 \rangle$ respectively that are used to encode a precedence constraint. For each precedence constraint, say $i \prec j$, a set of clauses that encode the propositional constraint $bp_{ij} \Rightarrow (\neg i_1 \wedge j_1) \vee ((i_1 \Leftrightarrow j_1) \wedge (\neg i_2 \vee j_2))$ are also added. The formulae associated with $j \prec k$ and $k \prec i$ are encoded similarly. Although $bp_{ij}$ and $bp_{jk}$ are set to true, UP does not infer $\neg bp_{ik}$, since none of the clauses obtained by applying Tseitin transformation is unary. Therefore, unlike AE, SE$_b$ does not detect the inconsistency.

Thus, we can infer that if unit propagation detects inconsistency in SE$_b$ then it also detects inconsistency in AE, but the converse is not true. $\square$

Given a set of assignments restricted to the problem variables, if unit propagation detects inconsistency in SE$_u$ then it also detects inconsistency in AE, and the converse is also true. This follows directly from the explanation of the symbol-based unary encoding and the atom-based encoding. Notice that both encodings detect cycles consisting of two features of the form $i \prec j$ and $j \prec i$. If the cycles involve more than two features $i \prec j$, $j \prec k$, $k \prec i$ both of them will infer $i \prec k$ which will result in a cycle consisting of two features $i$ and $k$.

## 10. Mixed Integer Linear Programming

In linear programming the goal is to optimise an objective function subject to linear equality and inequality constraints. When some variables are forced to be integer-valued, the problem is called Mixed Integer Linear Programming (MIP) problem. The standard way of expressing these problems is by presenting the function to be optimised, the linear constraints to be respected and the domain of the variables involved. Both the basic COP formulation and the atom-based PWMSAT formulation for finding an optimal relaxation of a feature subscription $\langle F, H, P, w \rangle$ can be translated into a MIP formulation. The translation of the PWMSAT formulation into MIP is straightforward. For this particular formulation we observed that CPLEX was not able to solve even simple problems within a time limit of 4 hours. In this paper, we only present the MIP formulation that corresponds to the basic COP formulation as presented in Section 2.2.

---

4. When there are only 2 features, the clauses corresponding to $[\![i \prec j]\!]$ in SE$_b$ are unary, in which case inconsistency can be detected by UP if it exists. However, the same inconsistency will be detected in the atom-based encoding.

**Variables.**    For each $i \in F$, we use a binary variable $bf_i$ and a real variable $pf_i$. A binary variable $bf_i$ is equal to 1 or 0 depending on whether feature $i$ is included or not. A real variable $pf_i$, $1 \leq pf_i \leq |F|$, if $bf_i$ is set to 1, is used to determine the position of the feature $i$ in the computed subscription. For each user precedence constraint $(i \prec j) \in P$, we use a binary variable $bp_{ij}$. It is instantiated to 1 or 0 depending on whether the precedence constraint $i \prec j$ holds or not.

**Linear Inequalities.**    If the features $i$ and $j$ are included in the computed subscription and if $(i \prec j) \in H$ then the position of feature $i$ must be less than the position of feature $j$. To this effect, we need to translate the underlying implication $(bf_i \wedge bf_j \Rightarrow (pf_i < pf_j))$ into the following linear inequality:

$$pf_i - pf_j + n * bf_i + n * bf_j \leq 2n - 1 \ . \tag{6}$$

Here, $n$ is a constant that is equal to the number of features, $|F|$, selected by the user. When both $bf_i$ and $bf_j$ are 1, Inequality (6) will force $(pf_i < pf_j)$. Note that this is not required for any user precedence constraint $(i \prec j) \in P$, since it can be violated.

A user precedence $(i \prec j) \in P$ is equivalent to the implication $bp_{ij} \Rightarrow (pf_i < pf_j) \wedge bf_i \wedge bf_j$, which in turn is equivalent to the conjunction of the three implications $(bp_{ij} \Rightarrow (pf_i < pf_j))$, $(bp_{ij} \Rightarrow bf_i)$ and $(bp_{ij} \Rightarrow bf_j)$. These implications can be translated into the following inequalities:

$$pf_i - pf_j + n * bp_{ij} \leq n - 1 \tag{7}$$

$$bp_{ij} - bf_i \leq 0 \tag{8}$$

$$bp_{ij} - bf_j \leq 0 \ . \tag{9}$$

Inequality (7) means that $bp_{ij} = 1$ forces $pf_i < pf_j$ to be true. Also, if $bp_{ij} = 1$ then both $bf_i$ and $bf_j$ are equal to 1 from Inequalities (8) and (9) respectively.

**Objective Function.**    The objective is to find an optimal relaxation of a feature subscription configuration problem $\langle F, H, P, w \rangle$ that maximises the sum of the weights of the features and the user precedence constraints that are selected:

$$\text{Maximise} \sum_{i \in F} w(i) \times bf_i + \sum_{(i \prec j) \in P} w(i \prec j) \times bp_{ij}.$$

## 11. Experimental Results

In this section, we shall describe the empirical evaluation of finding an optimal relaxation of randomly generated feature subscriptions using constraint programming, partial weighted maximum Boolean satisfiability and integer linear programming.

## 11.1 Problem Generation and Experimental Settings

In order to compare the different approaches we generated and experimented with a variety of *random catalogues* and many classes of *random feature subscriptions*. All the random selections below are performed with uniform distributions. A random catalogue is defined by a tuple $\langle f_c, B_c, T_c \rangle$. Here, $f_c$ is the number of features, $B_c$ is the number of binary constraints and $T_c \subseteq \{\prec, \succ, \prec\!\!\succ\}$ is a set of types of constraints. Note that $i \prec\!\!\succ j$ means that in any given subscription both features $i$ and $j$ cannot exist together. A random catalogue is generated by selecting $B_c$ pairs of features randomly from $f_c(f_c - 1)/2$ pairs of features. Each selected pair of features is then associated with a type of constraint that is selected randomly from $T_c$. A random feature subscription is defined by a tuple $\langle f_u, p_u, w \rangle$. Here, $f_u$ is the number of features that are selected randomly from $f_c$ features, $p_u$ is the number of user precedence constraints between the pairs of features that are selected randomly from $f_u(f_u - 1)/2$ pairs of features, and $w$ is an integer greater than 0. Each feature and each user precedence constraint is associated with an integer weight that is selected randomly between 1 and $w$ inclusive.

We generated catalogues of the following forms: $\langle 50, 250, \{\prec, \succ\} \rangle$, $\langle 50, 500, \{\prec, \succ, \prec\!\!\succ\} \rangle$ and $\langle 50, 750, \{\prec, \succ\} \rangle$. For each random catalogue, we generated classes of feature subscriptions of the following forms: $\langle 10, 5, 4 \rangle$, $\langle 15, 20, 4 \rangle$, $\langle 20, 10, 4 \rangle$, $\langle 25, 40, 4 \rangle$, $\langle 30, 20, 4 \rangle$, $\langle 35, 35, 4 \rangle$, $\langle 40, 40, 4 \rangle$, $\langle 45, 90, 4 \rangle$ and $\langle 50, 5, 4 \rangle$. Note that $\langle 50, 250, \{\prec, \succ\} \rangle$ is the default catalogue and the value of $w$ is 4 by default, unless stated otherwise. For each catalogue 10 instances of feature subscriptions were generated and their mean results are reported in the paper[5]. We remark that only 4 randomly generated instances were consistent out of the 270 generated instances. These consistent instances are instances of the feature subscription class $\langle 10, 5, 4 \rangle$ of catalogue $\langle 50, 250, \{\prec, \succ\} \rangle$.

All the experiments were performed on a PC Pentium 4 (CPU 1.8 GHz and 768MB of RAM) processor. The performances of all the approaches are measured in terms of search nodes (#nodes) and runtime in seconds (time). The time reported is the time spent in both finding the optimal solution and proving optimality. We used the time limit of 14,400 seconds (i.e., 4 hours) to cut the search. No initial bounds were computed for any of the approaches.

## 11.2 Evaluation of Constraint Programming Formulations

For the basic constraint optimisation problem model as presented in Section 7 we first investigated the effect of Maintaining Arc Consistency (MAC) within branch and bound search. We also studied the effect of maintaining different levels of consistency on different sets of variables within a problem. In particular we investigated, (1) maintaining singleton arc consistency on the Boolean variables and MAC on the remaining variables (see Section 7.2), and (2) maintaining restricted singleton arc consistency on the Boolean variables and MAC on the remaining variables (see Section 7.3); the former is denoted by MSAC$_b$ and the latter by MRSAC$_b$. All the branch and bound search algorithms were tested with two different variable ordering heuristics: *dom/deg* (Bessiere & Regin, 1996) and *dom/wdeg* (Boussemart, Hemery, Lecoutre, & Sais, 2004). Here *dom* is the domain size, *deg* is the original degree of a variable, and *wdeg* is the weighted degree of a variable. All the experiments for the basic constraint optimisation problem formulation were done using CHOCO[6] (version 2.1) a Java library for constraint programming systems. Some results for all the three branch and bound search algorithms with the *dom/deg* variable ordering heuristic are presented in Table 2 and with the *dom/wdeg* variable ordering heuristic are presented in Table 3.

---

5. All the generated instances are available on `http://4c.ucc.ie/~lquesada/FeatureSubscription/page/instances.htm`.

Table 2: Average results of MAC, MRSAC$_b$ and MSAC$_b$ with *dom/deg* heuristic.

| $\langle f_u, p_u \rangle$ | MAC | | MRSAC$_b$ | | MSAC$_b$ | |
|---|---|---|---|---|---|---|
| | time | #nodes | time | #nodes | time | #nodes |
| $\langle 20, 10 \rangle$ | 0.2 | 1,691 | **0.0** | 45 | **0.0** | 44 |
| $\langle 25, 40 \rangle$ | 9.8 | 70,233 | **0.5** | 174 | 0.6 | 156 |
| $\langle 30, 20 \rangle$ | 5.6 | 29,076 | **0.6** | 179 | 0.7 | 157 |
| $\langle 35, 35 \rangle$ | 125.2 | 479,650 | **7.3** | 1,269 | 8.1 | 1,083 |
| $\langle 40, 40 \rangle$ | 1,716.9 | 5,307,530 | **68.8** | 9,830 | 75.1 | 8,466 |

Table 3: Average results of MAC, MRSAC$_b$ and MSAC$_b$ with *dom/wdeg* heuristic.

| $\langle f_u, p_u \rangle$ | MAC | | MRSAC$_b$ | | MSAC$_b$ | |
|---|---|---|---|---|---|---|
| | time | #nodes | time | #nodes | time | #nodes |
| $\langle 20, 10 \rangle$ | 0.1 | 701 | **0.0** | 42 | **0.0** | 41 |
| $\langle 25, 40 \rangle$ | 3.3 | 20,096 | **0.5** | 164 | 0.6 | 145 |
| $\langle 30, 20 \rangle$ | 2.4 | 10,511 | **0.5** | 161 | 0.6 | 142 |
| $\langle 35, 35 \rangle$ | 76.9 | 248,447 | **5.5** | 932 | 6.3 | 798 |
| $\langle 40, 40 \rangle$ | 889.0 | 2,255,713 | **45.9** | 6,105 | 52.9 | 5,184 |

Tables 2 and 3 clearly show that maintaining (R)SAC on the Boolean variables and AC on the integer variables dominates maintaining AC on all the variables. To the best of our knowledge this is the first time that such a significant improvement has been observed by maintaining a partial form of singleton arc consistency during search. As the problem size increases the difference in terms of the number of nodes visited by MRSAC$_b$ and MSAC$_b$ increases. Note that MRSAC$_b$ usually visits more nodes than those visited by MSAC$_b$, but the difference between them is not that significant. This suggests that the level of consistency enforced by RSAC on the instances of feature subscription problem is very close to that enforced by SAC. Despite visiting more nodes, MRSAC$_b$ usually requires less time than MSAC$_b$. On average, all the three search algorithms perform better with the *dom/wdeg* heuristic than with the *dom/deg* heuristic. Note that in the remainder of the paper the results that correspond to the basic COP model are obtained using MRSAC$_b$ with the *dom/wdeg* variable ordering heuristic.

We remark that the underlying algorithms in MAC and MRSAC$_b$ that enforce AC and RSAC$_b$ respectively have an optimal worst-case time complexity. However, the underlying algorithm of MSAC$_b$ that enforces SAC$_b$ does not have an optimal worst-case time complexity. Implementing an algorithm to enforce SAC$_b$ that has an optimal worst-case time complexity is not only cumbersome but also has a higher space requirement. The works of Bessiere et al. (2004, 2005) provide evidence that when an optimal algorithm for enforcing SAC is used as a preprocessor it is very expensive both in terms of running time and space. Therefore, maintaining it during search, as in our case, could be even more expensive. Indeed there exists other sub-optimal but efficient algorithms for enforcing singleton arc consistency on constraints networks, as proposed by Lecoutre et al. (2005) and, it remains to see whether any of these efficient algorithms can reduce the running time of MSAC$_b$.

---

6. http://choco.sourceforge.net/

Notice that SAC$_b$ can prune more values than RSAC$_b$. However, in practice, the difference between their pruning on the instances of feature subscriptions is not much, as is evident based on the number of nodes and time shown in Tables 2 and 3. We recall that RSAC$_b$ enforces partial SAC$_b$. At a given node in the search tree, RSAC$_b$ enforces arc consistency at most one time for each assignment of a value to each Boolean variable, whereas SAC$_b$ can enforce arc consistency at most $n$ times in the worst-case. Here $n$ is the sum of the Boolean variables associated with features and user precedences. Nevertheless, in practice, we observed that it was much less. For example, for any instance of feature subscription of the class $\langle 40, 40 \rangle$ arc consistency was enforced at most 7 times for any variable-value pair, which is much less than $n = 80$. This also justifies the use of a non-optimal version of algorithm to enforce SAC$_b$.

Our WCSP formulation for finding an optimal relaxation of a feature subscription was also tested. For this purpose toulbar2 (a generic solver for WCSP) was used[7]. In general the results in terms of time were poor. We remark that a solution of the WCSP model is a total order on the features whose position variables are assigned values greater than 0. Due to holes (when a feature is excluded) different assignments of the position variables may lead to the same total order. Thus, more search effort could be spent for the WCSP formulation. We recall that in the basic COP model the decision variables are only the Boolean variables that indicate the inclusion/exclusion of features and user precedences and not the position variables. Therefore, an optimal solution of the basic COP model may not necessarily be a total order on the included features. Nevertheless, it can be obtained by computing a topological sort on the included user precedences and the catalogue precedences defined over the included features.

In order to remove the symmetries the WCSP formulation, as described in Section 8.2, can be augmented. One way could be to associate costs with the values (greater than 0) of the position variables in such a way that there is a unique assignment of values to the variables, which is optimal for a given strict partial order. Our preliminary investigation suggested that the number of nodes were reduced but at the expense of increasing the time. In our current setting, the WCSP approach has been used as a black box. Indeed, certain improvements can be made which may improve the performance in terms of time. For example, stronger soft consistency techniques can be applied similar to the singleton arc consistency for the COP model, which is more efficient for feature subscription problem.

We also investigated the impact of using the global constraint SoftPrec. This global constraint was implemented in CHOCO. The results obtained by using it are denoted by SP. Five variants of SoftPrec have been investigated by Lesaint et al. (2009). The results presented in this paper correspond to the variant that was observed to be the best in terms of time, which Lesaint et al. (2009) denoted by SP$_4$. The results in Tables 6-8 show that SoftPrec always outperforms MRSAC$_b$ on average. However, Lesaint et al. (2008a) theoretically showed that the pruning achieved by maintaining RSAC on the Boolean variables of the COP model and AC on the remaining variables is incomparable with the pruning achieved by using SOFTPREC.

---

7. http://carlit.toulouse.inra.fr/cgi-bin/awki.cgi/ToolBarIntro

### 11.3 Evaluation of the Boolean Satisfiability Formulations

The evaluation of the atom-based PWMSAT encoding of feature subscription was carried out on three different solvers: (a) SAT4J[8] (version 2.1.1), an efficient library of SAT solvers in Java that implements the MINISAT specification (Eén & Sörensson, 2003); (b) MINISAT+[9] (version 1.13+), a pseudo-Boolean solver implemented on top of MINISAT (Eén & Sörensson, 2006); and (c) CLASP[10] (version 1.3.0), an answer set solver that supports Boolean constraint solving (Gebser, Kaufmann, & Schaub, 2009). As the two last solvers are pseudo-Boolean solvers, the PWMSAT instances were translated into linear pseudo-Boolean instances by associating each clause with a linear pseudo-Boolean constraint, and defining the objective function as the weighted sum of the soft clauses in the PWMSAT model (de Givry, Larrosa, Meseguer, & Schiex, 2003).

The results of the evaluation are summarized in Table 4. We remark that the results for the SAT4J solver, especially for the dense catalogues, are roughly 10 times faster in terms of time when compared to those presented by Lesaint et al. (2008c). This is simply due to the advances in the version of the SAT4J that has been used to obtain the results. Despite that, SAT4J is significantly outperformed by both MINISAT+ and CLASP. We observed up to a one order-of-magnitude gap in those cases where the catalogue is sparse. CLASP and MINISAT+ seem to be incomparable in our instances. Even though CLASP performed better on our toughest category of instances $\langle 45, 90 \rangle$, CLASP spent 27% more time solving the whole set of instances. We also noticed that CLASP seems to be more sensitive to the number of features in sparse instances. While we observed a gap of one order-of-magnitude between categories $\langle 45, 90 \rangle$ and $\langle 50, 4 \rangle$ in the $\langle 50, 250, \{\prec, \succ\}\rangle$ catalogue with SAT4J and MINISAT+, the gap observed with CLASP was not that significant.

Table 4: Results for the atom-based encoding using different SAT solvers.

| | $\langle 50, 250, \{\prec, \succ\}\rangle$ | | | $\langle 50, 500, \{\prec, \succ, \prec\succ\}\rangle$ | | | $\langle 50, 750, \{\prec, \succ\}\rangle$ | | |
|---|---|---|---|---|---|---|---|---|---|
| $\langle f, p \rangle$ | SAT4J | CLASP | MINISAT+ | SAT4J | CLASP | MINISAT+ | SAT4J | CLASP | MINISAT+ |
| $\langle 30, 20 \rangle$ | 0.6 | **0.1** | 1.2 | 0.5 | **0.0** | 0.7 | 0.8 | **0.2** | 0.7 |
| $\langle 35, 35 \rangle$ | 2.7 | **0.8** | 3.0 | 0.7 | **0.1** | 1.3 | 2.5 | **0.8** | 2.0 |
| $\langle 40, 40 \rangle$ | 18.2 | **6.9** | 8.0 | 1.2 | **0.1** | 2.0 | 8.0 | **3.2** | 4.5 |
| $\langle 45, 90 \rangle$ | 1,156.4 | **111.1** | 119.6 | 3.6 | **0.4** | 5.7 | 46.7 | **13.8** | 25.5 |
| $\langle 50, 4 \rangle$ | 90.8 | **79.0** | 11.9 | 3.7 | **0.6** | 3.8 | 147.1 | 43.8 | **12.8** |

We now compare the atom-based encoding with the symbol-based unary and binary encodings as described in Section 9.2. In order to do a fair comparison between these encodings we need to solve the same instances of feature subscription on the same machine using the same solver. As we did not have access to the instances of feature subscription for $SE_u$ and $SE_b$ encodings, we use the results of the experiments run by Daniel Le Berre[11] for all the three encodings: AE, $SE_u$ and $SE_b$ on the same instances of feature subscription using SAT4J solver (version 2.1.0) on a PC Pentium 4 (CPU 3 GHz). Codish et al. (2009) have also made these results public.

---

8. http://www.sat4j.org/
9. http://minisat.se/MiniSat+.html
10. http://www.cs.uni-potsdam.de/clasp/
11. http://www.cril.univ-artois.fr/~leberre/

Table 5 presents results for feature subscriptions of different sizes of different catalogues for three encodings: AE, SE$_u$, and SE$_b$. The experimental results show that AE is, in general, more efficient than SE$_b$, which is consistent with the fact that unit propagation on AE is strictly stronger than unit propagation on SE$_b$. Note that AE is up to two orders-of-magnitude faster than SE$_b$. Notice that SE$_b$ never outperforms both SE$_u$ and AE on any class of feature subscription.

Table 5: Mean results in terms of time obtained using AE, SE$_u$, SE$_b$ encodings in SAT4J.

| subscription | $\langle 50, 250, \{\prec, \succ\}\rangle$ | | | $\langle 50, 500, \{\prec, \succ, \prec\succ\}\rangle$ | | | $\langle 50, 750, \{\prec, \succ\}\rangle$ | | |
|---|---|---|---|---|---|---|---|---|---|
| | AE | SE$_u$ | SE$_b$ | AE | SE$_u$ | SE$_b$ | AE | SE$_u$ | SE$_b$ |
| $\langle 10, 5\rangle$ | **0.05** | 0.12 | 0.17 | **0.07** | 0.29 | 0.15 | **0.07** | 0.31 | 0.16 |
| $\langle 15, 20\rangle$ | **0.12** | 0.69 | 0.32 | **0.13** | 0.95 | 0.30 | **0.14** | 1.12 | 0.47 |
| $\langle 20, 10\rangle$ | **0.15** | 0.76 | 0.36 | **0.17** | 1.24 | 0.42 | **0.18** | 1.32 | 0.79 |
| $\langle 25, 40\rangle$ | **0.41** | 1.70 | 1.87 | **0.27** | 1.75 | 1.23 | **0.35** | 2.44 | 5.90 |
| $\langle 30, 20\rangle$ | **0.58** | 1.66 | 1.22 | **0.31** | 2.21 | 1.49 | **0.47** | 3.58 | 9.16 |
| $\langle 35, 35\rangle$ | **1.40** | 3.46 | 7.12 | **0.57** | 3.15 | 3.35 | **1.33** | 7.19 | 49.65 |
| $\langle 40, 40\rangle$ | 9.20 | **9.06** | 21.03 | **0.91** | 3.73 | 5.31 | **3.22** | 15.67 | 153.75 |
| $\langle 45, 90\rangle$ | 484.16 | **161.37** | 1,844.01 | **2.34** | 8.85 | 22.11 | **24.64** | 64.79 | 1205.12 |
| $\langle 50, 4\rangle$ | 30.72 | **7.09** | 11.97 | **2.39** | 4.91 | 8.77 | 61.57 | **41.87** | 618.66 |

Although the results reported in Tables 1, 2 and 3 of the works of Codish et al. (2008, 2009) suggest that SE$_b$ is much better than AE, the results shown in Table 5 contradict this conclusion. The results obtained by using SE$_b$ are significantly outperformed by those obtained by using AE. This apparent conflict could be for one of several reasons. The results reported by Codish et al. (2008) were based on different instances for different encodings and the instances used for the symbol-based encoding were very much easier and in fact some large size instances with 50 features were already consistent. Also, the experiments for different encodings were conducted on different machines. Codish et al. (2008, 2009) obtained the results for the symbol-based encoding and the atom-based encodings using different solvers. The experiments for SE$_b$ were done using a solver, which has been implemented on top of MINISAT, while for AE the results were obtained using the SAT4J solver. It is apparent from Table 4 that the use of different solvers can make a huge difference in terms of runtime. In fact, we have observed a huge improvement for AE when tested with the MINISAT+ solver. This latter fact suggests that the speed up observed by Codish et al. (2008, 2009) could be mostly because of the use of MINISAT. Also, notice that the results depicted in Table 5 are in accordance with the fact that unit propagation in the atom-based encoding is strictly stronger than unit propagation in the symbol-based binary encoding.

Although unit propagation on AE encoding is equivalent to unit propagation on SE$_u$ encoding when assignments are restricted to problem variables, empirically it is not always possible to observe this due to the exploration of the search trees in different orders. Table 5 shows that AE and SE$_u$ are incomparable in terms of time. Therefore, it is not possible to conclude superiority of any of the two approaches. We have also been informed that the instances of the symbol-based encodings also include the computation of the objective function, and the comparison of the value of the objective function with an upper bound as described by Codish et al. (2009). However, they are not needed when applying the PWMSAT solver of SAT4J. These extra clauses may indeed prevent the symbol-based approaches to perform at their best. Nevertheless, most of the clauses of the symbol-based encodings are coming from the encoding of the precedence constraints.

Finding an optimal relaxation of a feature subscription using a SAT solver can be decomposed into three tasks: (a) the encoding of the strict partial order, (b) the encoding of the objective function, and (c) the underlying search algorithm of the SAT solver. Improving any of these tasks can improve the whole approach for solving the problem. In this paper we have focused on task (a), which is mainly about the encoding of the precedence constraints. We remark that (a), (b) and (c) are orthogonal tasks, so any of the techniques for tasks (b) and (c) can certainly be used with any of the techniques for task (a). The different encodings of precedence constraints can be fairly compared when the same (or the best suited) techniques of tasks (b) and (c) are used. Codish et al. (2008, 2009) propose several techniques for (b) and (c), e.g., the encoding of the sum constraint and the use of dichotomic search for the optimisation aspect. It may be possible to improve the results of atom-based encoding further by using these techniques.

## 11.4 Comparison between CP, SAT and MIP-based approaches

The performances of using constraint programming (CP), partial weighted maximum satisfiability (SAT) and mixed integer linear programming (MIP) approaches are presented in Tables 6, 7 and 8. The MIP model of the problem was solved using ILOG CPLEX[12] (version 10.1). For the CP approaches the results are presented for MRSAC$_b$ and the global constraint denoted by SP. For the SAT approaches we use the results obtained by using CLASP and MINISAT+. All the approaches solved all the instances within the time limit. Since in general finding an optimal relaxation is NP-hard, we need to investigate which approach can do it in reasonable time. The best approach in terms of time is represented in bold letters for each class of feature subscription.

Table 6: Catalogue $\langle 50, 250, \{\prec, \succ\}\rangle$.

| $\langle f_u, p_u\rangle$ | MIP | | CP | | | | SAT | | | |
| | | | MRSAC$_b$ | | SP | | CLASP | | MINISAT+ | |
| | #nodes | time | #nodes | time | #nodes | time | #nodes | time | #nodes | time |
| --- | --- | --- | --- | --- | --- | --- | --- | --- | --- | --- |
| $\langle 30, 20\rangle$ | 208 | 0.4 | 161 | 0.5 | 115 | 0.2 | 5,258 | **0.1** | 3,938 | 1.2 |
| $\langle 35, 35\rangle$ | 905 | 2.0 | 932 | 5.6 | 744 | 2.8 | 11,565 | **0.8** | 9,757 | 3.0 |
| $\langle 40, 40\rangle$ | 2,616 | 9.1 | 6,105 | 45.9 | 2,707 | 12.3 | 37,331 | **6.9** | 20,368 | 8.0 |
| $\langle 45, 90\rangle$ | 9,818 | **77.4** | 104,789 | 1,256.1 | 103,065 | 971.3 | 310,595 | 111.1 | 133,303 | 119.6 |
| $\langle 50, 4\rangle$ | 1,754 | **6.1** | 26,494 | 218.1 | 9,133 | 36.5 | 196,684 | 79.0 | 26,087 | 11.9 |

The results presented in Table 6 suggest that the MIP approach performs better than the CP and SAT approaches for the hardest feature subscription instances of the sparse catalogue $\langle 50, 250, \{\prec, \succ\}\rangle$, in particular for $\langle 45, 90\rangle$ and $\langle 50, 4\rangle$ classes of feature subscriptions, and for the remaining classes of feature subscription of the catalogue $\langle 50, 250, \{\prec, \succ\}\rangle$, the SAT approach based on the CLASP solver is the winner. For the dense catalogue $\langle 50, 750, \{\prec, \succ\}\rangle$, the MIP approach is significantly slower than the other approaches. Notice that the results for the MIP approach have improved significantly when compared with the results presented by Lesaint et al. (2008c). This is because of the usage of real-valued variables for the positions of features. The results presented in Tables 7 and 8 for the catalogues $\langle 50, 500, \{\prec, \succ, \prec\succ\}\rangle$ and $\langle 50, 750, \{\prec, \succ\}\rangle$, respectively, suggest that the SAT approaches perform significantly better than the MIP and CP approaches. In particular, the SAT approach based on the CLASP solver is the winner for all the classes except for the $\langle 50, 4 \rangle$ class of feature subscription of catalogue $\langle 50, 750, \{\prec, \succ\}\rangle$, where it is outperformed by the CP approach based on the global constraint and the SAT approach based on MINISAT+.

---

12. http://www.ilog.com/products/cplex/

Table 7: Catalogue $\langle 50, 500, \{\prec, \succ, \prec\succ\}\rangle$.

| | MIP | | CP | | | | SAT | | | |
|---|---|---|---|---|---|---|---|---|---|---|
| | | | MRSAC$_b$ | | SP | | CLASP | | MINISAT+ | |
| $\langle f_u, p_u \rangle$ | #nodes | time | #nodes | time | #nodes | time | #nodes | time | #nodes | time |
| $\langle 30, 20 \rangle$ | 48 | 0.4 | 66 | 0.2 | 53 | 0.1 | 2,066 | **0.0** | 4,298 | 0.7 |
| $\langle 35, 35 \rangle$ | 112 | 1.1 | 158 | 1.0 | 111 | 0.4 | 2,999 | **0.1** | 6,838 | 1.3 |
| $\langle 40, 40 \rangle$ | 160 | 1.8 | 229 | 1.8 | 188 | 1.0 | 4,005 | **0.1** | 8,897 | 2.0 |
| $\langle 45, 90 \rangle$ | 573 | 18.2 | 687 | 9.6 | 620 | 6.1 | 7,265 | **0.4** | 19,791 | 5.7 |
| $\langle 50, 4 \rangle$ | 258 | 1.5 | 768 | 6.2 | 954 | 3.6 | 8,887 | **0.6** | 16,511 | 3.8 |

Table 8: Catalogue $\langle 50, 750, \{\prec, \succ\}\rangle$.

| | MIP | | CP | | | | SAT | | | |
|---|---|---|---|---|---|---|---|---|---|---|
| | | | MRSAC$_b$ | | SP | | CLASP | | MINISAT+ | |
| $\langle f_u, p_u \rangle$ | #nodes | time | #nodes | time | #nodes | time | #nodes | time | #nodes | time |
| $\langle 30, 20 \rangle$ | 3,761 | 9.3 | 578 | 2.2 | 168 | 0.4 | 4,633 | 0.2 | 5,125 | 0.7 |
| $\langle 35, 35 \rangle$ | 13,485 | 67.9 | 1,997 | 11.4 | 396 | 1.9 | 9,285 | **0.8** | 12,611 | 2.0 |
| $\langle 40, 40 \rangle$ | 28,461 | 229.0 | 5,229 | 36.7 | 993 | 5.8 | 20,905 | **3.2** | 22,284 | 4.5 |
| $\langle 45, 90 \rangle$ | 43,958 | 539.1 | 19,190 | 207.8 | 2,902 | 29.7 | 60,676 | **13.8** | 60,531 | 25.5 |
| $\langle 50, 4 \rangle$ | 163,686 | 1,644.4 | 31,580 | 253.1 | 5,569 | 28.2 | 130,920 | 43.8 | 45,802 | **12.8** |

Even though MRSAC$_b$ and SOFTPREC are outperformed by at least one of the other approaches in all the cases, they are never the worst with respect to the total time required for solving all the instances as shown in Figure 3. In particular the CP approach based on SOFTPREC is very competitive in those cases where the catalog is dense. Figure 3 also shows that the pseudo-Boolean solvers CLASP and MINISAT+ perform better in terms of total time when compared with the other approaches. It should be noted that CLASP and MINISAT+ are implemented in C++ and use restarts, while MRSAC$_b$ and SOFTPREC are implemented in the Java-based CHOCO solver and they do not use restarts. Both CLASP and MINISAT+ perform poorly when compared with respect to the number of nodes visited during search. This shows that the time spent by CLASP and MINISAT+ at each node is considerably less than the time spent by the remaining approaches. There is of course the opportunity to improve the per-node speed of the CP approaches by implementing them in a C++ based solver. We also remark that both CLASP and MINISAT+ consume more memory than the CP-based approaches and the MIP approach. To illustrate this, we also computed the sum of the problem sizes of all the instances for all the approaches. Here, the problem size of an instance is the sum of the number of variables, the domain sizes of all the variables, and the arity of all the constraints. Figure 4 depicts the plot for the total problem size for each approach. The total problem size for CLASP and MINISAT+ is roughly two orders-of-magnitude more than the other approaches. We, therefore, conclude that CLASP and MINISAT+ do not offer scalability.

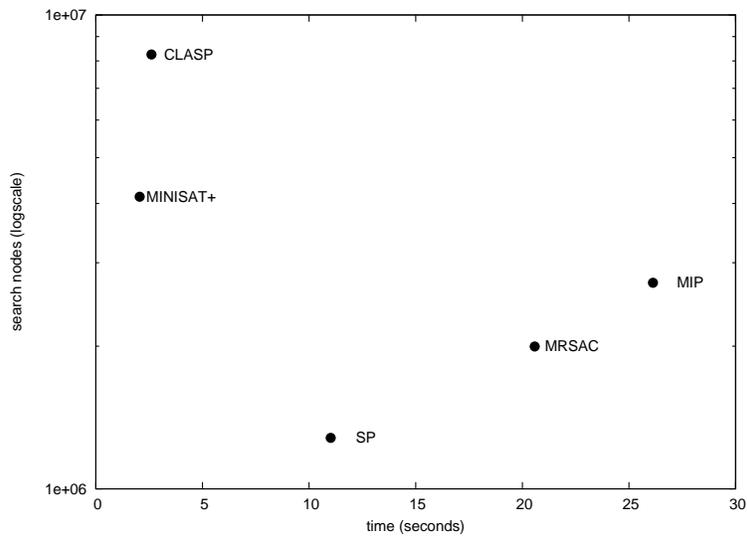

Figure 3: Total time and nodes required to solve all the instances by different approaches.

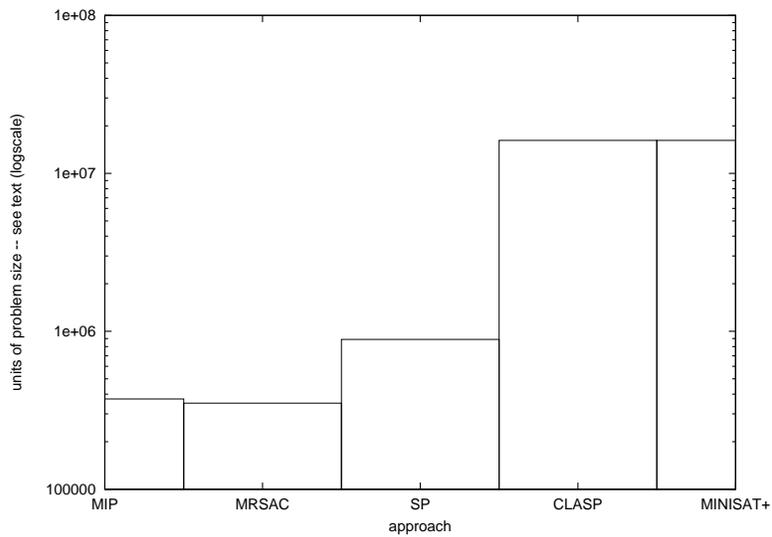

Figure 4: Total problem size of all the instances for different approaches.

## 12. Conclusions and Future Work

In this paper we have focussed on the task of finding an optimal relaxation of feature subscription when the user's preferences violate the technical constraints defined by a set of distributed feature composition rules. We reformulated the problem of finding an optimal relaxation, and showed that it is a generalisation of the Feedback Vertex Set problem, which makes the problem NP-hard. We developed CPbased methods for finding an optimal relaxation of feature subscription. In particular we presented three models: a basic constraint optimisation problem model, a model based on a global constraint, and a weighted CSP model. For the basic COP model, we studied the effect of maintaining arc consistency and two mixed consistencies during branch and bound search. Our experimental results suggest that maintaining (restricted) singleton arc consistency on the Boolean variables and arc consistency on the integer variables outperforms MAC significantly. The former approach was outperformed empirically by the CP approach based on the SOFTPREC global constraint.

We also compared the CPbased approaches with the SAT-based approaches and a mixed integer linear programming approach. In the partial weighted maximum satisfiability case we presented an atom-based encoding and investigated two symbol-based encodings. When the set of assignments are restricted to problem variables unit propagation on the atom-based encoding is strictly stronger than the unit propagation on the symbol-based binary encoding, and the former is equivalent to the unit propagation on the symbol-based unary encoding. Empirically, the atom-based encoding is better than the symbol-based binary encoding, and it is incomparable with the symbol-based unary encoding. Overall, the results suggest that when the catalogue is sparse MIP is better in terms of runtime on hard instances. When the catalogue is dense the SAT approach based on CLASP is better in terms of runtime. The SAT approach based on MINISAT+ and the CP approach based on the global constraint are also very competitive on the dense catalogues. Overall, the pseudo-Boolean solvers CLASP and MINISAT+ perform better in terms of total time when compared with the other approaches.

The approaches considered in this paper are mostly one-stage approaches in the sense that the exploration is started without any approximation of the optimum value. In the future we would like to consider a two-stage approach where, at the first stage, a heuristic is used to compute an approximation of the optimal solution, and at the second stage, the exploration is carried out taking the approximate value as an initial lower bound. The CP approach based on WCSP was explored the least. It may be possible to improve its performance by using different models that overcome the problem of symmetric solutions and stronger consistency techniques similar to singleton arc consistency in the case of the basic COP model. In the current settings the performance of all the approaches in terms of time includes the time taken to prove the optimality of the solution. In the future, we would like to compare all the presented approaches and also local search methods in terms of their anytime profiles (i.e. solution qualities over time). It would be interesting to investigate the impact of restarts on all the approaches.

## Acknowledgments

This material is based upon work supported by the Science Foundation Ireland under Grant No. 05/IN/I886, 08/PI/I1912, and Embark Post Doctoral Fellowships No. CT1080049908 and No. CT1080049909. The authors would like to thank Hadrien Cambazard, Daniel Le Berre and Alan Holland for their support in using CHOCO, SAT4J and CPLEX respectively. The authors would also like to thank Simon de Givry for his help in the WCSP formulation of the problem. Thanks also to Michael Codish for providing the symbol-based encoding instances to Daniel Le Berre. We thank all the reviewers for providing valuable comments that helped us to improve the quality of the paper.